\pdfoutput=1

\documentclass[11pt]{article}

\usepackage[backref=page]{hyperref}       

\usepackage{ACL2023}

\usepackage{makecell}

\usepackage{times}
\usepackage{latexsym}

\usepackage[T1]{fontenc}

\usepackage[utf8]{inputenc}

\usepackage{microtype}

\usepackage{inconsolata}

\usepackage[pdftex]{graphicx}
\usepackage{subcaption}
\usepackage{multirow}
\usepackage[inline]{enumitem}
\usepackage{amsthm,amsmath,amssymb}
\usepackage{array}

\usepackage{booktabs}       

\usepackage{url}            
\usepackage{makecell}
\usepackage{cleveref}

\usepackage{color, colortbl}
\newcommand{\mycc}{\cellcolor{gray!20}}
\newcommand{\mycctwenty}{\cellcolor{gray!55}}
\newcolumntype{M}[1]{>{\centering\arraybackslash}m{#1}}
\usepackage{makecell}
\usepackage{multirow}
\usepackage{xspace}

\newlength{\Oldarrayrulewidth}

\usepackage{placeins}

\newcommand{\VarCell}[2]{ \multicolumn{1}{#1}{\scriptsize $\pm$ #2} }

\usepackage{amsmath,amsfonts,bm}

\def\eqref#1{equation~\ref{#1}}

\def\1{\bm{1}}

\def\rmW{{\mathbf{W}}}

\def\vzero{{\bm{0}}}

\def\vc{{\bm{c}}}

\def\vh{{\bm{h}}}

\def\vq{{\bm{q}}}

\DeclareMathAlphabet{\mathsfit}{\encodingdefault}{\sfdefault}{m}{sl}
\SetMathAlphabet{\mathsfit}{bold}{\encodingdefault}{\sfdefault}{bx}{n}

\newcommand{\name}{{\fontfamily{bch}\selectfont Multi-CLS BERT}\xspace}

\newcommand{\NameNoSpace}{{\fontfamily{bch}\selectfont Multi-CLS BERT}}

\title{\name: An Efficient Alternative to Traditional Ensembling}

\author{
Haw-Shiuan Chang\thanks{\, indicates equal contribution} \thanks{\, The work is done while the authors were at UMass} \, \, \, Ruei-Yao Sun\footnotemark[1] \footnotemark[2]  \, \, \,  \\ Amazon \\ USA \\ \texttt{chawshiu@amazon.com} \\ \texttt{rueiyas@amazon.com}   
\And 
Kathryn Ricci\footnotemark[1] \, \, \,  Andrew McCallum \\ 
  CICS, UMass, Amherst\\
  140 Governors Dr., Amherst, MA, USA\\
  \texttt{kathryn.d.ricci@gmail.com} \\ \texttt{mccallum@cs.umass.edu}
 \\}

\begin{document}
\maketitle
\begin{abstract}
Ensembling BERT models often significantly improves accuracy, but at the cost of significantly more computation and memory footprint. In this work, we propose \name, a novel ensembling method for CLS-based prediction tasks that is almost as efficient as a single BERT model. \name uses multiple CLS tokens with a parameterization and objective that encourages their diversity. Thus instead of fine-tuning each BERT model in an ensemble (and running them all at test time), we need only fine-tune our single \name model (and run the one model at test time, ensembling just the multiple final CLS embeddings).  To test its effectiveness, we build \name on top of a state-of-the-art pretraining method for BERT~\citep{aroca2020losses}.  In experiments on GLUE and SuperGLUE we show that our \name reliably improves both overall accuracy and confidence estimation. When only 100 training samples are available in GLUE, the \NameNoSpace$_{\text{Base}}$ model can even outperform the corresponding BERT$_{\text{Large}}$ model.  We analyze the behavior of our \name, showing that it has many of the same characteristics and behavior as a typical BERT 5-way ensemble, but with nearly 4-times less computation and memory.

\end{abstract}

\section{Introduction}

BERT (Bidirectional Encoder Representations from
Transformers)~\citep{BERT} is one of the most widely-used language model (LM) architectures for natural language understanding (NLU) tasks. We often fine-tune the pretrained BERT or its variants such as RoBERTa~\citep{liu2019roberta} 
so that the LMs learn to aggregate all the contextualized word embeddings into a single CLS embedding 
for a downstream text classification task.

\begin{figure}[t!]
\centering
\includegraphics[width=1\linewidth]{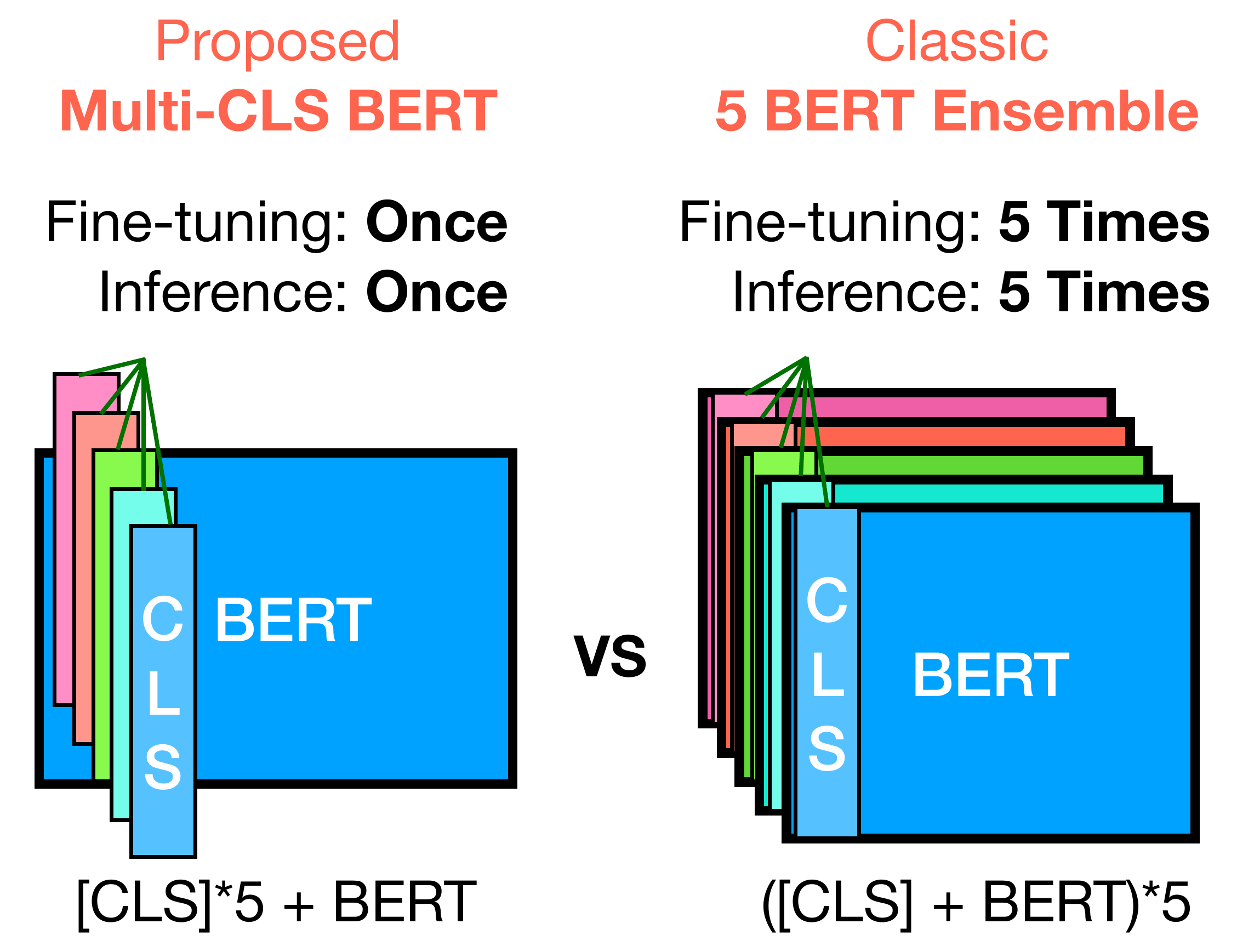}
\caption{Comparison of \name and the classic BERT ensemble. \name only ensembles the multiple CLS embeddings in one BERT encoder rather than ensemble multiple BERT encoders with different parameter weights.}
\label{fig:first_page}
\end{figure}

During fine-tuning, different initializations and different training data orders significantly affect BERT's generalization performance, especially with a small training dataset \citep{dodge2020fine,zhang2020revisiting,mosbach2021stability}. 
One simple and popular solution to the issue is to fine-tune BERT model multiple times using different random seeds and ensemble their predictions to improve its accuracy and confidence estimation. Although very effective, the memory and computational cost of ensembling a large LM is often prohibitive~\citep{xu2020improving, liang2022camero}. Naturally, we would like to ask, ``Is it possible to ensemble BERT models at no extra cost?''

To answer the question, we propose \name, which enjoys the benefits of ensembling without sacrificing efficiency. Specifically, we input the multiple CLS tokens to BERT and encourage the different CLS embeddings to 
aggregate the information from different aspects of the input text. As shown in Figure~\ref{fig:first_page}, the proposed \name shares all 
the hidden states of the input text and only ensembles different ways of aggregating the hidden states. Since the input text is usually much longer than the number of inputted CLS embeddings, \name is almost as efficient as the original BERT.

\citet{abs-2012-09816} discovered that the key of an effective ensembling model is the diversity of individual models and the models trained using different random seeds have more diverse predictions compared to simply using dropout~\citep{srivastava2014dropout,gal2016dropout} or averaging the weights of the models during training~\citep{abs-1912-02757}. To ensure the diversity of CLS embeddings without fine-tuning \name using multiple seeds, we propose several novel diversification techniques. For example, we insert different linear layers into the transformer encoder for different CLS tokens. Furthermore, we propose a novel re-parametrization trick to prevent the linear layers from learning the same weights during fine-tuning. 

We test the effectiveness of these techniques by modifying the multi-task pretraining method proposed by \citet{aroca2020losses}, which combines four self-supervised losses. In our experiments, we demonstrate that the resulting \name can significantly improve the accuracy on GLUE~\citep{wang2018glue} and SuperGLUE~\citep{wang2019superglue}, especially when the training sizes are small. Similar to the BERT ensemble model, we further show that multiple CLS embeddings significantly reduce the expected calibration error, which measures the quality of prediction confidence, on the GLUE benchmark.

\subsection{Main Contributions}
\begin{itemize}[leftmargin=.2in,topsep=0pt]
\setlength\itemsep{0.0em}
    \item We propose an efficient ensemble BERT model that does not incur any extra computational cost other than inserting a few CLS tokens and linear layers into the BERT encoder.
    Furthermore, we develop several diversification techniques for pretraining and fine-tuning the proposed \name model.\footnote{We release our code at \url{https://github.com/iesl/multicls/}.}
    \item We improve the GLUE performance reported in \citet{aroca2020losses} using a better and more stable fine-tuning protocol and verify the effectiveness of its multi-task pretraining methods in GLUE and SuperGLUE with different training sizes.
    \item Building on the above state-of-the-art pretraining and fine-tuning for BERT, our experiments and analyses show that \name significantly outperforms the BERT due to its similarity to a BERT ensemble model. The comprehensive ablation studies confirm the effectiveness of our diversification techniques.
\end{itemize}

\section{Method}

In sections \ref{sec:multi_pretraining} and \ref{sec:QT_loss}, we first review its state-of-the-art pretraining method from \citet{aroca2020losses}. In \Cref{sec:multi_QT_loss}, we modify one of its losses, quick thoughts (QT), to pretrain our multiple embedding representation. 
In \Cref{sec:contrastive_hard_neg}, we encourage the CLS embeddings to capture the fine-grained semantic meaning of the input sequence by adding hard negatives during the pretraining. To diversify the CLS embeddings, we modify the transformer encoder in \Cref{sec:architecture} and propose a new reparametrization method during the fine-tuning 
in \Cref{sec:contrastive_FT}.

\begin{figure*}[t!]
\centering
\includegraphics[width=0.98\linewidth]{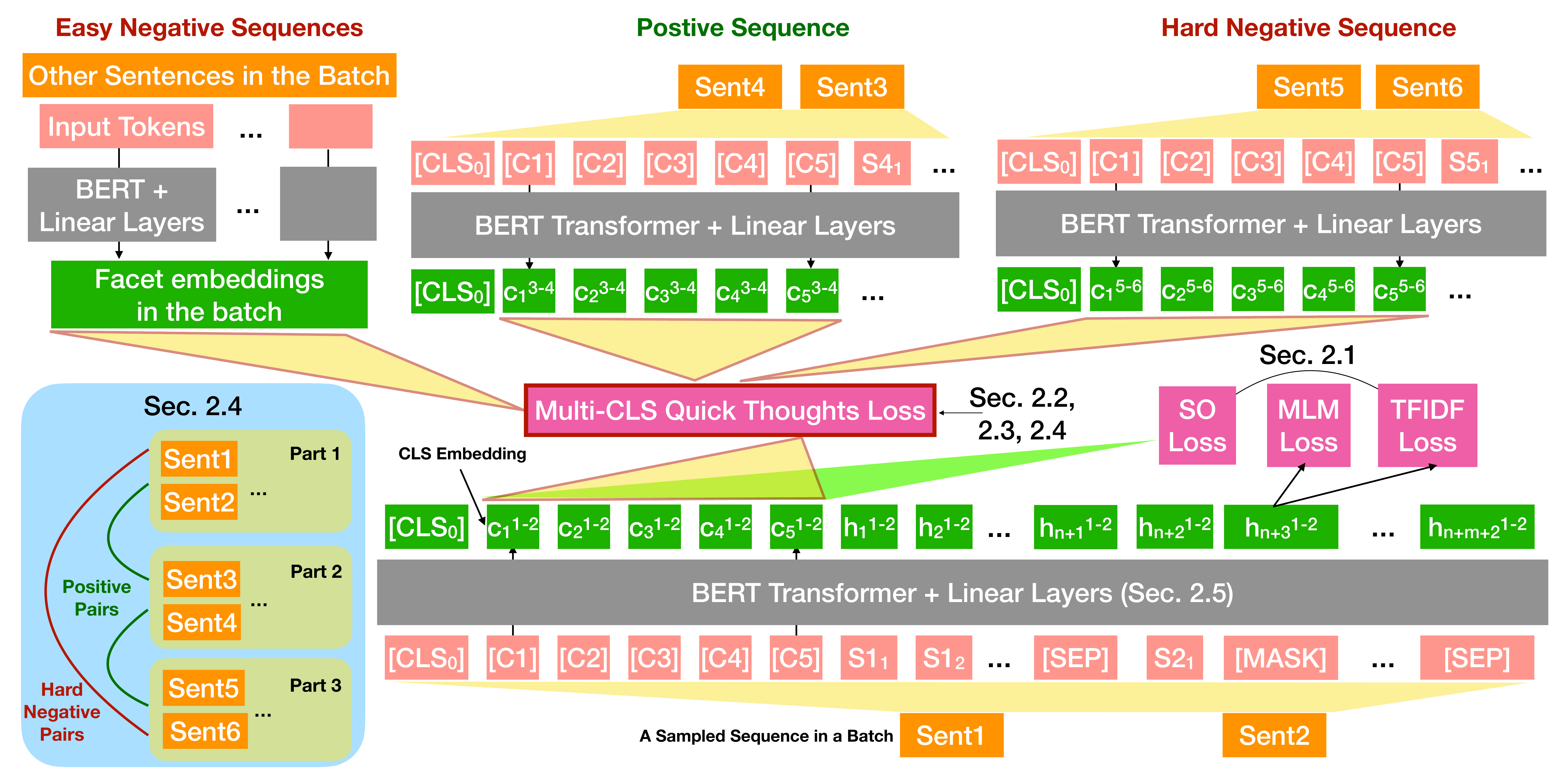}
\caption{Our MCQT, SO, MLM, and TFIDF loss, which are a modification of multi-task pretraining proposed in \citet{aroca2020losses}. 
The multi-CLS quick thought (MCQT) loss maximizes the CLS similarities between a sequence (sentences 1 and 2) and the next sequence (sentences 3 and 4) while minimizing the CLS similarities to other random sequences and the sequence after the next one (sentences 5 and 6). Notice that sentence 4 is inputted before sentence 3 because the sentence order is swapped for the SO loss.}
\label{fig:contrastive_loss}
\end{figure*}

\subsection{Multi-task Pretraining}
\label{sec:multi_pretraining}

After testing many self-supervised losses, \citet{aroca2020losses} find that combining the masked language modeling (MLM) loss, TFIDF loss, sentence ordering (SO) loss~\citep{sun2020ernie}, and quick thoughts (QT) loss \citep{LogeswaranL18} could lead to the best performance. 

The MLM loss is to predict the masked words and the TFIDF loss is to predict the importance of the words in the document. Each input text sequence consists of multiple sentences. They swap the sentence orders in some input sentences and use the CLS embedding to predict whether the order is swapped in the SO loss. Finally, QT loss is used to encourage the CLS embeddings of the consecutive sequences to be similar.

To improve the state-of-the-art pretraining method, we modify the multi-task pretraining method by using multiple CLS embeddings to represent the input sequence and using non-immediate consecutive sentences as the hard negative. Our training method is illustrated in \autoref{fig:contrastive_loss}. 


\subsection{Quick Thoughts Loss}
\label{sec:QT_loss}

Two similar sentences tend to have the same label in a downstream application, so pretraining should pull the CLS embeddings of these similar sentences closer. The QT loss achieves this goal by assuming consecutive text sequences are similar and encouraging their CLS embeddings to be similar.

\citet{aroca2020losses} propose an efficient way of computing QT loss in a batch by evenly splitting each batch with size $B$ into two parts. The first part contains $B/2$ text sequences randomly sampled from the pretrained corpus, and
the second part contains each of the $B/2$ sentences that are immediately subsequent to those in the first part. 
Then, for each sequence in the first part, they use the consecutive sequence in the second part as the positive example and the other $B/2-1$ sequences as the negative examples. We can write the QT loss for the sequences containing sentences 1, 2, 3, and 4 as 

\vspace{-3mm}
\small
\begin{equation}
\label{eq:logit_cross_entropy}
L_{QT}(s^{1-2}, s^{3-4}) = - \log(\frac{ \exp(\text{Logit}^{QT}_{s^{1-2},s^{3-4}}) }{\sum_{s} \exp(\text{Logit}^{QT}_{s^{1-2},{s}})}),
\end{equation}
\normalsize
where $s$ is the sentences in the second part of the batch, $\text{Logit}^{QT}_{s^{1-2},s^{3-4}} = (\frac{\vc^{{1-2}}}{||\vc^{{1-2}}||})^T \frac{\vc^{{3-4}}}{||\vc^{{3-4}}||}$ is the score for classifying sequence $s^{3-4}$ as the positive example, $\frac{\vc^{{1-2}}}{||\vc^{{1-2}}||}$ is the L2-normalized CLS embedding for sentences 1 and 2. 
The normalization is intended to stabilize the pretraining by limiting the gradients' magnitudes.

\subsection{Multiple CLS Embeddings}
\label{sec:multi_QT_loss}

A text sequence could have multiple facets; two sequences could be similar in some facets but dissimilar in others, especially when the text sequences are long. The QT loss squeezes all facets of a sequence into a single embedding and encourages all facets of two consecutive sequences to be similar, potentially causing information loss. 

Some facets might better align with the goal of a downstream application. For example, the facets that contain more sentiment information would be more useful for sentiment analysis. To preserve the diverse facet information during pretraining, we propose multi-CLS quick thoughts loss (MCQT). 
The loss integrates two ways of computing the similarity of two sequences. The first way computes the cosine similarity between the most similar facets, and the second computes the cosine similarity between the summations of all facets. We linearly combine the two methods as the logit of the two input sequences:

\vspace{-3mm}
\small
\begin{align}
\label{eq:logit_multi}
\text{Logit}^{MC}_{s^{1-2},s^{3-4}} = \lambda  \max_{i,j} (\frac{\vc_i^{{1-2}}}{||\vc_i^{{1-2}}||})^T \frac{\vc_j^{{3-4}}}{||\vc_j^{{3-4}}||} + \nonumber \\
(1-\lambda) (\frac{\sum_i \vc_i^{{1-2}}}{||\sum_i \vc_i^{{1-2}}||})^T \frac{\sum_j \vc_j^{{3-4}}}{||\sum_j \vc_j^{{3-4}}||}.
\end{align}
\normalsize
where $\lambda$ is a constant hyperparameters; $\vc^{1-2}_k$ and $\vc^{3-4}_k$ are the CLS embeddings of sentences 1-2 and sentences 3-4, respectively.  

The first term only considers the most similar facets to allow some facets to be dissimilar. 
Furthermore, the term implicitly diversifies CLS embeddings by considering each CLS embedding independently.
In contrast, the second term encourages the CLS embeddings to work collaboratively, as in a typical ensemble model, and also let every CLS embedding receive gradients more evenly. 
Notice that we sum the CLS embeddings before the normalization so that the encoder could predict the magnitude of each CLS embedding as its weight in the summation.

To show that the proposed method can improve the state-of-the-art pretraining methods, we keep the MLM loss and TFIDF loss unchanged. For the sentence ordering (SO) loss, we project the $K$ hidden states $\vh^c_k$ into the embedding $\vh^{SO}$ with the hidden state size $D$ for predicting the sentence order: $\vh^{SO} = L^{SO}(\oplus_k \vh^c_k )$, where $\oplus_k \vh^c_k$ is the concatenation of $K$ hidden states with size $K \times D$.

\subsection{Hard Negative}
\label{sec:contrastive_hard_neg}

For a large transformer-based LM, distinguishing the next sequence from random sequences could be easy. The LM can achieve low QT loss by outputting nearly identical CLS embeddings for the sentences with the same topic while ignoring the fine-grained semantic information~\citep{papyan2020prevalence}. In this case, multiple CLS embeddings might become underutilized. 

The hard negative is a common method of adjusting the difficulties of the contrastive learning~\citep{soares2019matching, cohan2020specter}.
Our way of collecting hard examples is illustrated in the bottom-left block of \autoref{fig:contrastive_loss}. To efficiently add the hard negatives in the pretraining, we split the batch into three parts. For each sequence in the first part, we would use its immediate next sequence in the second part as the positive example, use the sequence after the next one in the third part as the hard negative, and use all the other sequences in the second or the third part as the easy negatives. We select such sequence after the next one as our hard negatives because 
the sequence usually share the same topic with the input sequence but is more likely to have different fine-grained semantic facets compared to the immediate next sequence. 

After adding the hard negative, the modified QT loss of the three consecutive sequences becomes 

\vspace{-3mm}
\small
\begin{align}
\label{eq:logit_ce_hard}
L_{MCQT}(s^{1-2}, s^{3-4}, s^{5-6}) = \nonumber \\ 
- \log \left( \frac{ \exp(\text{Logit}^{MC}_{s^{1-2},s^{3-4}}) }{  \sum\limits_{s \in \{s^{3-4}, ..., s^{5-6}, ...\}  } \exp(\text{Logit}^{MC}_{s^{1-2},{s}})} \right)  \nonumber \\
- \log \left( \frac{ \exp(\text{Logit}^{MC}_{s^{5-6},s^{3-4}}) }{  \sum\limits_{s \in \{s^{3-4}, ..., s^{1-2}, ...\} } \exp(\text{Logit}^{MC}_{s^{5-6},{s}})} \right), 
\end{align}
\normalsize
where MCQT refer to multi-CLS quick thoughts, $\{s^{3-4}, ..., s^{5-6}, ...\}$ are all the sequences in the second and the third part, and $\{s^{3-4}, ..., s^{1-2}, ...\}$ are all the sequences in the first and the second part.

\begin{figure}[t!]
\centering
\includegraphics[width=1\linewidth]{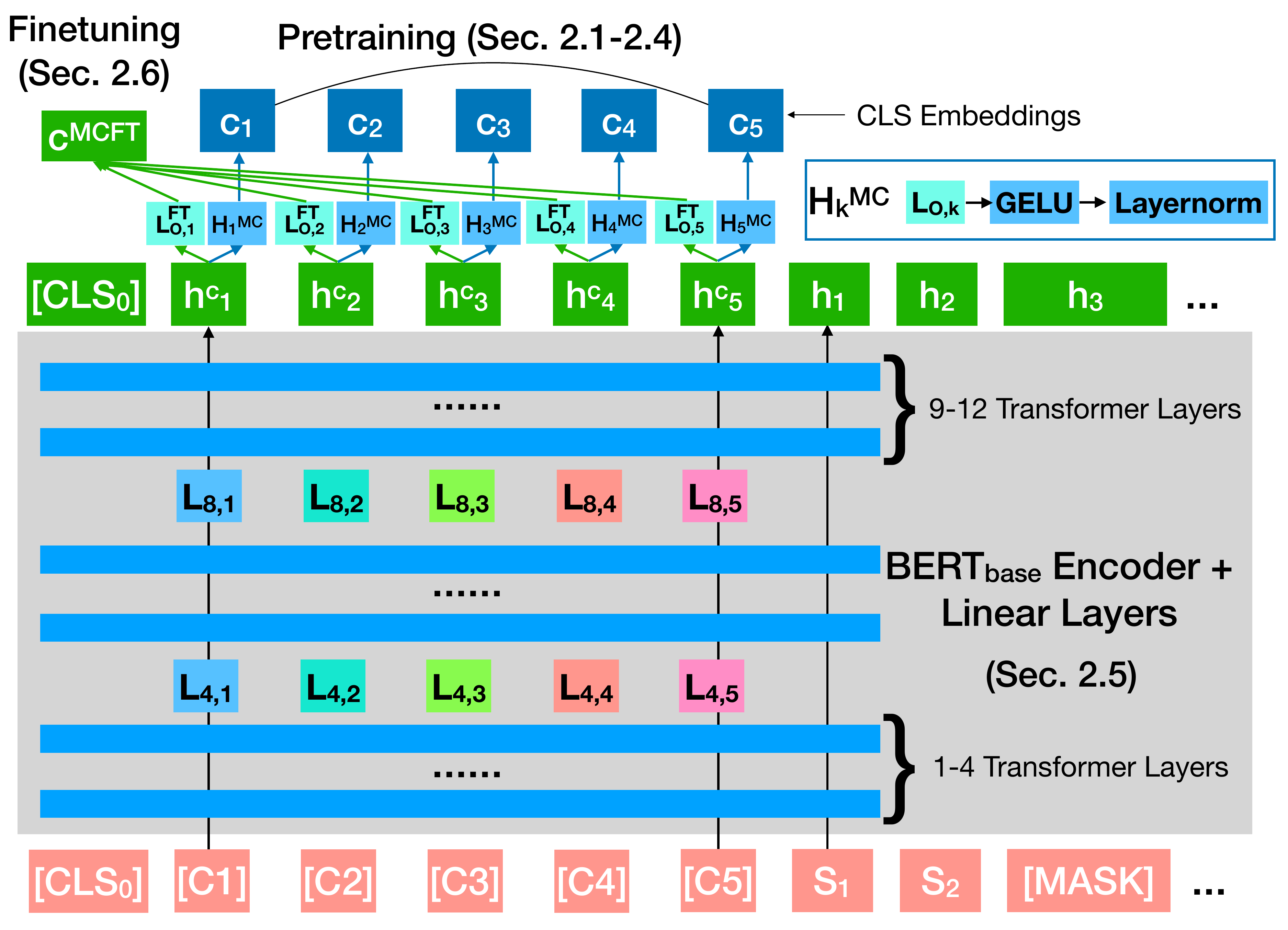}
\caption{The architecture of \name encoder that is built on BERT$_{\text{Base}}$ model. 
The different linear layers are applied to the hidden states corresponding to different CLS tokens
to increase the diversity of the resulting CLS embeddings. }
\label{fig:contrastive_architecture}
\end{figure}






\subsection{Architecture-based Diversification}
\label{sec:architecture}

Initially, we simply input multiple special CLS tokens ([C1], ..., [CK]) after the original CLS token, [CLS$_0$], and take the corresponding hidden states as the CLS embeddings, but we found that the CLS embeddings quickly become almost identical during the pretraining. 

Subsequently, instead of using the same final transformation head $H^{QT}$ for all CLS hidden states, we use a different linear layer $L_{O,k}$ in the final head $H_k^{MC}$ to transform the hidden state $\vh^c_k$ for the $k$th CLS. We set the bias term in $L_{O,k}$ to be the constant $\vzero$ because we want the differences between the CLS to be dynamic and context-dependent.
 
Nevertheless, even though we differentiate the resulting CLS embeddings $\vc_k = H_k^{MC}(\vh^c_k)$, the hidden states $\vh^c_k$ before the transformation usually still collapse into almost identical embeddings.

To solve the collapsing problem, we insert multiple linear layers $L_{l,k}$ into the transformer encoder. In \autoref{fig:contrastive_architecture}, we illustrate our encoder architecture built on the BERT$_{\text{Base}}$ model. 
After the $4$th transformer layer, we insert the layers $L_{4,k}$ to transform the hidden states before inputting them to the $5$th layer. Similarly, we insert $L_{8,k}$ between the $8$th transformer layer and $9$th transformer layer. For BERT$_{\text{Large}}$, we insert $L_{l,k}(.)$ after layer 8 and layer 16. Notice that although the architecture looks similar to the adapter~\citep{houlsby2019parameter} or prefix-tuning~\citep{li2021prefix}, our purpose is to diversify the CLS embeddings rather than freezing parameters to save computational time.

\subsection{Fine-Tuning}
\label{sec:contrastive_FT}

As shown in \autoref{fig:contrastive_architecture}, we input multiple CLS tokens into the BERT encoder during fine-tuning and pool the corresponding CLS hidden states into the single CLS embedding for each downstream task fine-tuning in order to avoid overfitting and increasing computational overhead. As a result, we can use the same classifier architecture on top of \name and BERT, which also simplifies their comparison. 

We discover that simply summing all the CLS hidden states still usually makes the hidden states and the inserted linear layers (e.g., $L_{O,k}$) almost identical after fine-tuning.
To avoid collapsing, we aggregate the CLS hidden states by proposing a novel re-parameterization trick: 
\begin{equation}
\label{eq:L_out_ft}
\vc^{MCFT} 
= \sum_k \left( L^{FT}_{O,k}(\vh^c_k) \right), 
\end{equation}
where $L^{FT}_{O,k}(\vh^c_k) = (\rmW_{O,k} - \frac{1}{K} \sum_{k'} \rmW_{O,k'})\vh^c_k$, and $\rmW_{O,k}$ is the linear weights of $L_{O,k}$. Then, if all the $L^{FT}_{O,k}$ become identical (i.e., $\forall k, \rmW_{O,k}=\frac{1}{K} \sum_{k'} \rmW_{O,k'}$),  $L^{FT}_{O,k}(\vh^c_k) = \vzero = \vc^{MCFT}$. However, gradient descent would not allow the model to constantly output the zero vector, so $L^{FT}_{O,k}$ remains different during the fine-tuning.

\section{Experiments}

The parameters of neural networks are more restricted as more training samples are available~\citep{mackay1995probable} and the improvement of deep ensemble models comes from the diversity of individual models~\citep{abs-1912-02757}, so the benefits of ensembling are usually more obvious when the training set size is smaller. Therefore, in addition to using the full training dataset, we also test the settings where the models are trained by 1k samples~\citep{zhang2020revisiting} or 100 samples from each task in GLUE~\citep{wang2018glue} or SuperGLUE~\citep{wang2019superglue}. Another benefit of the 1k- and 100-sampling settings is that the average scores would be significantly influenced by most datasets rather than by only a subset of relatively small datasets~\citep{card2020little}. 

\subsection{Experiment Setup}
To accelerate the pretraining experiments, we initialize the weights using the pretrained BERT models~\citep{BERT} and continue the pretraining using different loss functions on Wikipedia 2021 and BookCorpus~\cite{Zhu_2015_ICCV}. 

All of the methods are based on uncased BERT as in \citet{aroca2020losses}. We compare the following methods:
\begin{itemize}[leftmargin=.1in,topsep=0pt]
\setlength\itemsep{-0.2em}
    \item \textbf{Pretrained}: The pretrained BERT model released from \citet{BERT}.
    \item \textbf{MTL}: Pretraining using the four losses selected in \citet{aroca2020losses}: MLM, QT, SO, and TFIDF. We remove the continue learning procedure used in ERNIE~\citep{sun2020ernie} because we find that simply summing all the losses leads to better performance (see our ablation study in \Cref{sec:contrastive_ablation}). 
    \item \textbf{Ours (K=5, $\mathbf{\lambda}$)}: The proposed \name method using 5 CLS tokens. We show the results of setting $\lambda=\{0,0.1,0.5,1\}$ in \autoref{eq:logit_multi}. We reduce the maximal sentence length by 5 to accommodate the extra 5 CLS tokens.
    \item \textbf{Ours (K=1)}: 
    We set $K=1$ in our method to verify the effectiveness of using multiple embeddings. During fine-tuning, the CLS embedding is a linear transformation of the single facet $\text{CLS} = L_{O,1}(\vh^f_1)$.
\end{itemize}

\begin{table*}[t!]
\scalebox{0.85}{
\begin{tabular}{ccc|ccc|ccc|}
                            &                  &  & \multicolumn{3}{c|}{GLUE}                               & \multicolumn{3}{c|}{SuperGLUE}     \\
                         Configuration $\downarrow$   & Model Name $\downarrow$ & Model Size $\downarrow$ & 100             & 1k               & Full              & 100*             & 1k*               & Full              \\ \hline
\multirowcell{10}{BERT \\ Base}  & Pretrained  & 109.5M      & 55.71          & 71.67          & 82.05          & 57.18          & 61.55          & 65.04 \vspace{-0.2cm}         \\
& & & \VarCell{r}{0.62}   &	\VarCell{r}{0.15} &	\VarCell{r|}{0.08} &	\VarCell{r}{0.43} &	\VarCell{r}{0.37} \vspace{-0.1cm} &	\VarCell{r|}{0.36}   \\	
                      & MTL   & 109.5M & 59.29          & 73.26          & 83.30$\dagger$          & 57.50          & 62.94          & 66.33 \vspace{-0.2cm}          \\
                      & & & \VarCell{r}{0.27}   &	\VarCell{r}{0.13}   &	\VarCell{r|}{0.07}   &	\VarCell{r}{0.41}   &	\VarCell{r}{0.36}  \vspace{-0.1cm} &	\VarCell{r|}{0.33} \\
                      & Ours (K=1)   & 111.3M       & 57.84          & 73.28          & 83.40          & 57.31 & 63.35          & 66.29 \vspace{-0.2cm} \\
                      & & & \VarCell{r}{0.32}   &	\VarCell{r}{0.13}   &	\VarCell{r|}{0.07}   &	\VarCell{r}{0.35}   &	\VarCell{r}{0.18} \vspace{-0.1cm}  &	\VarCell{r|}{0.18} \\

                    & Ours (K=5, $\lambda=0$) & 118.4M      & 61.54          & \textbf{74.14}          & 83.41          & 58.29          & 63.71          & \textbf{66.80} \vspace{-0.2cm}         \\
                    & & & \VarCell{r}{0.32}   &	\VarCell{r}{0.12} &	\VarCell{r|}{0.07} &	\VarCell{r}{0.33} &	\VarCell{r}{0.26} \vspace{-0.1cm} &	\VarCell{r|}{0.25}   \\	
                    
                      & Ours (K=5, $\lambda=0.1$) & 118.4M & \textbf{61.80} &	74.10&	\textbf{83.47}&	58.20&	63.61 &	66.74 \vspace{-0.2cm} \\
                      & & & \VarCell{r}{0.35}   &	\VarCell{r}{0.13}   &	\VarCell{r|}{0.05}   &	\VarCell{r}{0.31}   & 	\VarCell{r}{0.27} \vspace{-0.1cm} &	\VarCell{r|}{0.26}\\
                      
                    & Ours (K=5, $\lambda=0.5$) & 118.4M       & 60.49          & 74.02          & \textbf{83.47}          & \textbf{58.41}          & \textbf{63.78}          & \textbf{66.80} \vspace{-0.2cm}         \\
                    & & & \VarCell{r}{0.35}   &	\VarCell{r}{0.12} &	\VarCell{r|}{0.08} &	\VarCell{r}{0.38} &	\VarCell{r}{0.25} \vspace{-0.1cm} &	\VarCell{r|}{0.24}   \\	
                    
                & Ours (K=5, $\lambda=1$) & 118.4M      & 59.86          & 73.75          & 83.43          & 57.84          & 63.56          & 66.39 \vspace{-0.2cm}         \\
                & & & \VarCell{r}{0.34}   &	\VarCell{r}{0.14} &	\VarCell{r|}{0.07} &	\VarCell{r}{0.40} &	\VarCell{r}{0.22}  &	\VarCell{r|}{0.22}   \\	
                      \hline

\multirowcell{9}{BERT \\ Large} &  
MTL & 335.2M & 61.39          & 75.30          & 84.13                & 59.03 & 65.21 & 69.16  \vspace{-0.2cm}     \\
& & & \VarCell{r}{0.37}   &	\VarCell{r}{0.27}   &	 \VarCell{r|}{0.11}  &	\VarCell{r}{0.54}   &	\VarCell{r}{0.38} \vspace{-0.1cm}   & \VarCell{r|}{0.37}	\\

                      & Ours (K=1)  & 338.3M     & 59.19          & 75.35          & 84.59                & 57.35          & 64.67          & 69.24  \vspace{-0.2cm}              \\
                      & & & \VarCell{r}{0.43}   &	\VarCell{r}{0.21}   &	\VarCell{r|}{0.07}   &	\VarCell{r}{0.42}   &	\VarCell{r}{0.43} \vspace{-0.1cm}	  & \VarCell{r|}{0.41}	\\
                      
                      &  Ours (K=5, $\lambda=0$)  & 350.9M & 63.19          & 75.73          & 84.51                & 59.46 & 65.43 & 69.56  \vspace{-0.2cm}     \\
& & & \VarCell{r}{0.49}   &	\VarCell{r}{0.26}   &	 \VarCell{r|}{0.05}  &	\VarCell{r}{0.44}   &	\VarCell{r}{0.38} \vspace{-0.1cm}   & \VarCell{r|}{0.31}	\\
                      
                      & Ours (K=5, $\lambda=0.1$)  & 350.9M    & \textbf{64.24} & \textbf{76.27} & \textbf{84.61}       & \textbf{59.88}          & 65.58          & \textbf{70.03}  \vspace{-0.2cm}    \\   
                      & & & \VarCell{r}{0.40}   &	\VarCell{r}{0.12}   &	\VarCell{r|}{0.08}   &	\VarCell{r}{0.43}   &	\VarCell{r}{0.26} \vspace{-0.1cm}   & \VarCell{r|}{0.25}	\\ 

                     &  Ours (K=5, $\lambda=0.5$)  & 350.9M & 63.02          & 75.95          & 84.49 & 59.42 & \textbf{65.84} & 69.79  \vspace{-0.2cm}     \\
& & & \VarCell{r}{0.42}   &	\VarCell{r}{0.10}   &	 \VarCell{r|}{0.08}  &	\VarCell{r}{0.34}   &	\VarCell{r}{0.25} \vspace{-0.1cm}   & \VarCell{r|}{0.25}	\\

                      &  Ours (K=5, $\lambda=1$)  & 350.9M & 62.07          & 75.85          & \textbf{84.61}                & 58.74 & 65.00 & 69.04  \vspace{-0.2cm}     \\
& & & \VarCell{r}{0.45}   &	\VarCell{r}{0.17}   &	 \VarCell{r|}{0.07}  &	\VarCell{r}{0.50}   &	\VarCell{r}{0.29}  & \VarCell{r|}{0.27}	\\ \hline
                      

\end{tabular}
}
\centering
\caption{The macro average scores on the development set. All numbers are percentages. The standard errors are shown as the confidence intervals. We make the best scores of the model built on BERT$_{\text{Base}}$ boldface and similar for the models built on BERT$_{\text{Large}}$. $\dagger$The number is much higher than $81.4$, the GLUE score reported by \citet{aroca2020losses} because we continue training from the pretrained BERT and we use better fine-tuning hyperparameters. *The scores do not contain ReCoRD in SuperGLUE.\footnotemark}
\label{tb:NLU_results}
\end{table*}

The GLUE and SuperGLUE scores are significantly influenced by the pretraining random seeds~\citep{BERT_robust} and fine-tuning random seeds~\citep{dodge2020fine,zhang2020revisiting,mosbach2021stability}. To stably evaluate the performance of different pretraining methods, we pretrain models using four random seeds and fine-tune each pretrained model using four random seeds, and report the average performance on the development set across all 16 random seeds. To further stabilize the fine-tuning process and reach better performance, we follow the fine-tuning suggestions from \citet{zhang2020revisiting} and \citet{mosbach2021stability}, including training longer, limiting the gradient norm, and using Adam~\cite{kingma2015adam} with bias term and warmup.

\subsection{Main Results}

Our results are presented in \autoref{tb:NLU_results}. 
We can see that \textbf{Ours (K=5)} is consistently better than other baselines and that the improvement is larger in datasets with fewer training samples. For example, in GLUE 100, it achieves $61.80$ on average using BERT$_{\text{Base}}$ with 118.4M parameters, which outperforms \textbf{MTL} using BERT$_{\text{Large}}$ with 335.2M parameters ($61.39$). Please see \Cref{sec:task_scores} for the scores of individual tasks. \textbf{MTL} significantly improves the scores of original BERT model (\textbf{Pretrained}), confirming the effectness of the QT, SO, and TFIDF losses. Compared to \textbf{MTL}, \textbf{Ours (K=1)} is slightly better in GLUE 1k and GLUE Full, but worse in GLUE 100.

We observe that $\lambda=0.1$ usually performs well, which justifies the inclusion of both the highest logit and average logit in \autoref{eq:logit_multi}. The $\lambda=0$ model has significantly worse performance only in BERT$_{\text{Large}}$ model. This suggests that the benefits of \name depend on our pretraining method and maximizing the highest logit stabilizes the pretraining of a larger model.

\footnotetext{In SuperGLUE 100 and 1k, we exclude the ReCoRD dataset because the performance of all models is much worse than the most frequent class baseline.}




\subsection{Ablation Study}
\label{sec:contrastive_ablation}

In our ablation studies, we would like to test the effectiveness of the design choices in our baseline
\textbf{MTL} and our best model, \textbf{Ours (K=5, $\mathbf{\lambda=0.1}$)}. The model variants we test include:
\begin{itemize}[leftmargin=.1in,topsep=0pt]
\setlength\itemsep{-0.2em}
    \item \textbf{MLM only}: Removing the QT, SO, and TFIDF losses in \textbf{MTL}. That is, we simply continue training \textbf{Pretrained} using only the MLM loss.
    \item \textbf{CMTL+}: The best pretrained method reported in \citet{aroca2020losses}. It uses the continual learning method~\citep{sun2020ernie} to weight each loss in \textbf{MTL}.
    \item \textbf{MLM+SO+TFIDF}: \textbf{MTL} without the QT loss.
    \item \textbf{No Inserted Layers}: Removing the $L_{l,k}(.)$ in the transformer encoder from our method.
    \item \textbf{No Hard Negative}: Removing the hard negatives described in \Cref{sec:contrastive_hard_neg} from our method.
    \item \textbf{Sum Aggregation}: Simply summing the facets (i.e., using $L_{O,k}$ to replace $L^{FT}_{O,k}$ in \autoref{eq:L_out_ft}). This baseline removes the proposed reparametrization trick to test its effectiveness. 
    \item \textbf{Default}: \textbf{Ours (K=k, $\mathbf{\lambda=0.1}$)}, where $k = \{1,3,5,10\}$.
    \item \textbf{SWA}: Stochastic weight averaging~\citep{ruppert1988efficient,izmailov2018averaging}  averages the weights along the optimization trajectory. 
    \item \textbf{Ensemble on Dropouts}: Running the forward pass of \textbf{Ours (K=1)} with dropout using 5 different seeds and averaging their prediction probabilities for each class in each task.
    \item \textbf{Ensemble on FT Seeds}: Fine-tuning \textbf{Ours (K=1)} or \textbf{Ours (K=5, $\mathbf{\lambda=0.1}$)} using 5 different seeds and averaging their prediction probabilities.
\end{itemize}


\begin{table}[t!]
\scalebox{0.62}{
\begin{tabular}{ccc|cc|cc|}
      &                    &    & \multicolumn{2}{c|}{GLUE}                   & \multicolumn{2}{c|}{SuperGLUE*}                  \\
    Model $\downarrow$  & Model Description $\downarrow$       & K $\downarrow$  & 100              & 1k             & 100                            & 1k             \\ \hline
\multirowcell{5}{Baselines \\ (BERT \\ Base)} & 
Pretrained & 1 & 56.85 & 71.68 & 57.90 & 62.14 \\
& MLM only & 1 & 55.38 & 70.74 & 57.39 & 61.77 \\
& CMTL+              & 1  & 58.65          & 72.57          & 56.88                        & 62.63          \\
  & MLM + SO + TFIDF         & 1  & 60.35 &	72.65 &	57.88 &	62.60	               \\
      & MTL    & 1  & 59.53          & 73.12          & 57.51                        & 62.95          \\ \hline
\multirowcell{11}{Ours \\ (BERT \\ Base)}       & No Inserted  & 1  & 58.06          & 73.18          & 57.97                        & 63.34          \\ 
      & Layers                   & 5  & 60.12          & 73.35          & 56.46                        & 62.00          \\ \cline{2-7}

& No Hard   & 1  & 58.44          & 73.30          & 57.19                        & 63.33          \\ 
      & Negative   & 5  & 61.77          & 74.18          & \textbf{58.89}                        & \textbf{63.86}          \\ \cline{2-7}
      
      & Sum Aggregation & 5  & 58.87          & 73.94          & 57.41                        & 63.82 \\ \cline{2-7}
      & \multirow{4}{*}{Default}               & 1  & 57.76          & 73.30          & 57.53                        & 63.22          \\
      &                    & 3  & 61.09          & 73.95          & 57.85                        & 63.31          \\
      &                    & 5  & \textbf{62.62}          & \textbf{74.49} & 58.82                        & \textbf{63.86}          \\
      &                    & 10 & 60.99          & 73.59          & 58.25               & 62.82          \\ \cline{2-7}
    & SWA & 1  &    57.31  &  72.91          & -                              & -                \\  \cline{2-7} \noalign{\vskip\doublerulesep
         \vskip-\arrayrulewidth}  \cline{2-7}
          & Ensemble on Dropouts & 1  &  58.45    &  72.86          & -                              & -                \\  \cline{2-7}
      & \multirow{2}{*}{Ensemble on FT Seeds}           & 1  & 60.07          & 75.20          & -                              & -                \\ 
      &            & 5  & 63.34          & 75.35          & -                              & -                \\ 
        \hline \hline
      
      \multirowcell{4}{Ours \\ (BERT \\ Large)}       & No Hard  & 1  & 60.36          & 75.69         & 58.47                        & 65.04          \\ 
      & Negative                   & 5  & 63.23          & 75.77          & \textbf{60.33}                        & \textbf{65.75}          \\ \cline{2-7}

    & \multirow{2}{*}{Default}  & 1  & 60.01          & 76.03         & 57.38                        & 65.10          \\ 
      &                    & 5  & \textbf{64.33}          & \textbf{76.38}          & 59.99                        & 65.51 \\ \hline \hline
\end{tabular}
}
\centering
\caption{The macro average scores on the development set for our ablation study. 
We highlight the best performance after excluding the ensemble baselines, which require much more computation. The scores are different in \autoref{tb:NLU_results} because we use two pretraining random seeds instead of four in the ablation study. SWA refers to Stochastic weight averaging~\citep{izmailov2018averaging}.
*SuperGLUE score does not contain ReCoRD. 
}
\label{tb:NLU_ablation}
\end{table}



Our results are presented in \autoref{tb:NLU_ablation}. We can see that continuing training using \textbf{MLM only} loss degrades the performance, which indicates that our improvement does not come from training BERT longer. Removing QT loss results in mixed results. The better performance of \textbf{MTL} compared to \textbf{CMTL+} suggests that the continual training technique used in \citet{aroca2020losses} is harmful with our evaluation settings. 

Removing the inserted layers (\textbf{No Inserted Layers}) or removing the re-parametrization trick (\textbf{Sum Aggregation}) makes the performance of \textbf{Ours (K=5, $\mathbf{\lambda=0.1}$)} close to the \textbf{Ours (K=1)} baseline. This result highlight the importance of diversity of CLS embeddings. The performance of \textbf{Ours (K=3)} and \textbf{Ours (K=10)} is usually better than \textbf{Ours (K=1)}, but are worse than \textbf{Ours (K=5)}. In both BERT$_{\text{Base}}$ and BERT$_{\text{Large}}$ models, removing hard negatives degrades the GLUE scores but slightly increases the SuperGLUE scores.

In GLUE 100 and 1k, we do not get good results by using other efficient ensembling methods such as \textbf{SWA} and \textbf{Ensemble on Dropouts}. This suggests that the gradient descent trajectory and different dropout maps might not produce prediction diversity sufficient for an effective BERT ensemble model~\citep{abs-1912-02757}.

On the other hand, ensembling the models that are fine-tuned using different random seeds indeed boosts the performance at the expense of high computational costs. The ensembled \name (\textbf{Ensemble on FT Seeds K=5}) still outperforms the ensembled \textbf{K=1} baseline, but ensembling makes their performance differences smaller. These results imply that the improvements of \name overlap with the improvements of a BERT ensemble model.








\begin{table}[t!]
\scalebox{0.8}{
\begin{tabular}{c|c|cc|}
& Inference & \multicolumn{2}{|c|}{GLUE* (ECE)} \\
& Time (s) & 100 & 1k\\ \hline
Ours (K=1) & 0.2918  &  25.22 & 19.32 \vspace{-0.2cm} \\
& \VarCell{r|}{0.0002} & \VarCell{r}{1.99} \vspace{-0.1cm}  &	\VarCell{r|}{1.64} \\
Ours (K=5, $\lambda=0.1$) & 0.3119 & \textbf{15.46} &  \textbf{17.01} \vspace{-0.2cm} \\ 
& \VarCell{r|}{0.0004} & \VarCell{r}{1.79} \vspace{-0.1cm}  &	\VarCell{r|}{1.64} \\ \hline \hline
Ensemble of Ours (K=1) & 1.4590 & 13.85 & 10.80 \vspace{-0.2cm} \\
 & \VarCell{r|}{0.0012} & \VarCell{r}{0.97}   &	\VarCell{r|}{0.88} \\ \hline

\end{tabular}
}
\centering
\caption{The comparison of inference time and expected calibration error (ECE). The confidence intervals are standard errors. *Only includes the classification tasks (i.e., excludes STS-b).
}
\label{tb:NLU_ECE}
\end{table}

\begin{table}[t!]
\scalebox{0.9}{
\begin{tabular}{c|cc|}
& GLUE* 100 & GLUE* 1k \\ \hline
Multi-CLS vs ENS &  32.57 & 41.35 \\
Dropout vs ENS & 37.17 & 45.53 \\ 
Least vs ENS & 39.57 & 48.85 \\
ENS vs ENS & 38.67 & 50.14 \\ \hline
\end{tabular}
}
\centering
\caption{The overlapping ratio of the top 20\% most uncertain examples using different uncertainty estimation methods. ENS is ensemble of Ours (K=5, $\lambda=0.1$) with different fine-tuning seeds. *Only includes the classification tasks (i.e., excludes STS-b).
}
\label{tb:NLU_top20}
\end{table}

\subsection{Ensembling Analysis}
We compare the inference time and expected calibration error (ECE)~\citep{naeini2015obtaining} of using multiple CLS embeddings, using a single CLS embedding, and ensembling BERT models with different fine-tuning seeds in \autoref{tb:NLU_ECE}. A lower ECE means a better class probability estimation. For example, if a model outputs class 1 with 0.9 probability for 100 samples, ECE = 0 means that 90 samples among them are indeed class 1. 

\autoref{tb:NLU_ECE} shows that \textbf{Ours (K=5)} is much faster than the BERT ensemble and almost as efficient as \textbf{Ours (K=1)}, because a BERT ensemble needs to run for multiple forward passes and we reduce the maximal sentence length by 5 in \textbf{Ours (K=5)}. Additionally, the ECE of \textbf{Ours (K=5)} is lower than \textbf{Ours (K=1)} but not as low as the ECE from ensembling BERT models with different fine-tuning seeds. That is, without significantly increasing inference time, ensembling multiple CLS embeddings improves the output confidence, even though not as much as ensembling BERT models.

Next, we analyze the correlation of uncertainty estimation from different methods in \autoref{tb:NLU_top20}.
When ensembling BERT models with different dropout maps (\textbf{Dropout}) or different fine-tuning seeds (\textbf{ENS}), we can estimate the prediction uncertainty by the variance of the prediction probability from each individual BERT model. We can also use one minus prediction probability as the uncertainty (\textbf{Least}). In \textbf{Multi-CLS}, we measure the disagreement among the CLS embeddings as the uncertainty\footnote{See \Cref{sec:ensembling_analysis_details} for details} and would like to see how many top-20\% most uncertain samples from the disagreement of CLS embeddings are also the top-20\% most uncertain samples for a BERT ensemble model.

\autoref{tb:NLU_top20} reports the ratio of the number of the overlapping uncertain samples from two estimation methods to the number of 20\% samples in the development set. We can see that the ratio from \name and the BERT ensemble model (\textbf{Multi-CLS vs ENS}) is close to the ratios from other uncertainty estimations and the BERT ensemble model (\textbf{Dropout vs ENS},  \textbf{Least vs ENS}, and \textbf{ENS vs ENS}). 
This shows that different CLS embeddings can classify the uncertain samples differently, as is the case for the different BERT models in a BERT ensemble model. 
In \Cref{sec:emb_vis}, we visualize the CLS embeddings of some uncertain samples to show how different CLS embeddings solve a task in different ways.




\section{Related Work}

Due to its effectiveness, ensembling BERT in a better or more efficient way has recently attracted researchers' attention. Nevertheless, the existing approaches often need to rely on distillation~\citep{xu2020improving, matsubara2022ensemble,zuo2022moebert} or still require significant extra computational cost during training and testing~\cite{kobayashi2022diverse, liang2022camero}.

Some recent vision models can also achieve ensembling almost without extra computational cost by sharing the weights~\citep{wen2019batchensemble}, partitioning the model into subnetworks~\citep{havasi2020training,zhang2021ex}, or partitioning the embeddings~\citep{lavoie2022simplicial}. However, it is unknown if the approaches are applicable to the pretraining and fine-tuning of language models.

Similar to \name, mixture of softmax (MoS) \citep{yang2018breaking} also uses multiple embeddings to improve the pretraining loss. Recently, \citet{narang2021transformer, tay2022scaling} have found that MoS is one of the few modifications that can improve on the original BERT architecture on the NLU benchmarks. Nevertheless, \citet{narang2021transformer} also point out that MoS requires significant extra training cost to compute the multiplication between each hidden state and all the word embeddings in the vocabulary.

\citet{chang2021extending} propose represent the sentence using multiple embeddings and demonstrate its improvement over the single embedding baseline on unsupervised sentence similarity tasks. Similar to our \autoref{eq:logit_multi}, their non-negative sparse coding loss also encourages multiple sentence embeddings to collaborate during pretraining. Nevertheless, our loss is more computationally efficient and is designed to improve downstream supervised tasks rather than similarity tasks. 

Some approaches also represent a text sequence using multiple embeddings, such as contextualized word embeddings~\citep{khattab2020colbert, luan2021sparse} for information retrieval applications, sentence embeddings~\citep{liu2019text, iter2020pretraining, mysore2021multi,sul2023balancing}, or entity pair embeddings~\citep{xue2022embarrassingly}. However, the goal of this approach is to improve the representation of a relatively long text sequence and it is unknown if its benefits could be extended to the GLUE tasks that require fine-tuning and often involve only one or two sentences.

\section{Conclusion}


In this work, we propose representing the input text using $K$ CLS embeddings rather than using the single CLS embedding in BERT. Compared to BERT, \name significantly increases the GLUE and SuperGLUE scores and reduces the expected calibration error in GLUE, while its only added cost is to reduce the maximal text length by $K$ and add
a little extra time for computing the inserted linear transformations.
Therefore, we recommend the wide use of multiple CLS embeddings for the almost free performance gain.

To solve the collapsing problem of CLS embeddings, we modify the pretraining loss, BERT architecture, and fine-tuning loss. The ablation study shows that all of these modifications contribute to the performance improvement of \name. 
In our analyses for investigating the source of the improvement, we find that 
\begin{enumerate*} [label=\alph*)]%
\item ensembling the original BERT leads to greater improvement than ensembling the \name and
\item the disagreement of different CLS embeddings highly correlates with the disagreement of the BERT models from different fine-tuning seeds.
\end{enumerate*}
Both findings support our perspective that \name is an efficient ensembling method.

\section{Acknowledgement}
We thank Jay Yoon Lee and the anonymous reviewers for their constructive feedback.
This work was supported 
in part by the Center for Data Science and the Center for Intelligent Information Retrieval, 
in part by the Chan Zuckerberg Initiative under the project Scientific Knowledge Base Construction, 
in part by the IBM Research AI through the AI Horizons Network, 
in part using high performance computing equipment obtained under a grant from the Collaborative R\&D Fund managed by the Massachusetts Technology Collaborative, 
and in part by the National Science Foundation (NSF) grant numbers IIS-1922090 and IIS-1763618.
Any opinions, findings, conclusions, or recommendations expressed in this material are those of the authors and do not necessarily reflect those of the sponsor.

\section{Limitations}

Our methods are evaluated using BERT as many previous recent work such as~\citet{aroca2020losses, dodge2020fine, BERT_robust, gu2021transformer, qin2021bibert, wang2021exploring, xu2022dense, zhou2022closer, hou2022token, wang2022skipbert, liu2022flooding, zhao2022fine, zhou2022bert, zheng2022robust, fu2022contextual}. Our limited computational resources do not allow us to conduct similar experiments on RoBERTa~\citep{liu2019roberta} because pretraining RoBERTa requires much powerful GPUs and a much larger CPU memory to store the corpora. For the similar reason,
we are unable to test our methods on larger language models. We are also not able to conduct a more comprehensive search for the pretraining and fine-tuning hyperparameters. We haven't tested if the multiple embedding representation could also improve other language model architectures such as XLNet~\citep{yang2019xlnet}, or 
other fine-tuning methods such as prompt~\citep{radford2019language,li2021prefix}, or adapter~\citep{houlsby2019parameter,wang2022adamix}.

Our conclusion mainly draws from the overall scores of GLUE or SuperGLUE benchmarks, which only include English datasets and might contain some dataset selection bias~\citep{dehghani2021benchmark}.
 
Although much more efficient, \name is still worse than the classic BERT ensemble model in terms of expected calibration error and accuracy when more training data are available (e.g., in GLUE 1k). We also do not know if \name could provide efficient and high-quality uncertainty estimation for other applications such as active learning~\cite{pop2018deep}.

\section{Ethical and Broader Impact}
\label{impact}

\name can provide better confidence estimation compared to BERT while better efficiency compared to the classic BERT ensemble. This work might inspire prospective efficient ensembling approaches that produce more robust predictions~\citep{clark2019don} with lower the energy consumption. 

On the other hand, the readers of the paper might not notice the limitations of the study (e.g., the confidence estimation of \name is still sometimes far behind the classic BERT ensemble model) and mistakenly believe that \name has all the benefits of the classic BERT ensemble model.

\bibliography{custom}
\bibliographystyle{acl_natbib}

\clearpage

\appendix

\section{Appendix Overview}
In the appendix, we first describe the details of our methods and evaluation protocol in \autoref{sec:exp_details}. Then, we visualize the disagreement of CLS embeddings of some samples in \autoref{sec:emb_vis} and provide a diversity metric during pretraining in \autoref{sec:diversity_metric}. 
Finally, we compare the performance of individual tasks in \autoref{sec:task_scores}.

\section{Experiment Details}
\label{sec:exp_details}

We first describe the architecture details and pretraining details of our methods and baselines in \Cref{sec:model_details}. Then, we list the hyperparameter setup in the fine-tuning in \Cref{sec:FT_details}. Finally, we explain the details of the ensemble baselines and their related analyses in
\Cref{sec:ensembling_analysis_details}. 

\subsection{Our Models and Baselines}
\label{sec:model_details}

The models built on BERT$_{\text{Base}}$ are pretrained using two billion tokens and each batch contains 30 sequences. The models built on BERT$_{\text{Large}}$ are pretrained using one billion tokens and each batch contains 48 sequences. The learning rate is $2 \cdot 10^{-5}$ and the warmup ratio is $0.001$ for the pretraining stage.

We implement \name by modifying the code of \citet{aroca2020losses}\footnote{\url{https://github.com/StephAO/olfmlm}}. We use [unused0] -- [unused(K-1)] tokens in the original BERT tokenizer as our input CLS tokens [C1] -- [CK]. We still keep the original CLS tokens to increase the comparability with the \textbf{MTL} baseline. 

We use NVIDIA GeForce RTX 2080, 1080, and TITAN X, M40 GPUs for the BERT$_{\text{Base}}$ experiments and use GeForce RTX 8000 and Tesla M40 for the BERT$_{\text{Large}}$ experiments. In \autoref{tb:NLU_results}, the model size excludes the top classifier parameters used in each task. 

We test \textbf{CMTL+} using the default hyperparameters of \citet{aroca2020losses} and we do not try different hyperparameters or different schedules of pretraining losses. \textbf{No Inserted Layers} only removes the $L_{l,k}(.)$ while still using different $H_k^{MC}$ on top during pretraining. \textbf{SWA} averages the weights of every model checkpoint that is evaluated using the validation dataset.








\subsection{Fine-tuning}
\label{sec:FT_details}

We start from the default evaluation hyperparameters used in \citet{aroca2020losses} and modify the settings based on the suggestions from \citet{zhang2020revisiting} and \citet{mosbach2021stability}. We find that the best hyperparameters depend on the training size. For example, batch size 16 works well in GLUE Full but is much worse than batch size 4 in GLUE 100. Furthermore, the performance of the default hyperparameters on some tasks is suboptimal or unstable even after averaging the performance from 16 trials. Therefore, we coarsely tune the hyperparameters to maximize and stabilize the performance of the \textbf{Ours (K=1)} baseline under the memory and computational time constraints in our GPUs. The preliminary results suggest that the hyperparameters also maximize the performance of \textbf{MTL}.

Next, we list fine-tuning hyperparameters for all the tasks\footnote{We use different values for some hyperparameters in ReCoRD. See the details below.}. Our fine-tuning stops after 20 epochs, 60k batches, or consecutive 10k batches without a validation improvement (whichever comes first). We use the first 5k validation samples to select the best fine-tuned model checkpoints for the evaluation. The maximal gradient norm is 1. The maximal length for sentences and CLS tokens is 128 for GLUE and 256 for SuperGLUE.

For each task, we select the best learning rate from $c \cdot 10^{-5}$ and $c=1,2,3,4,5,7$. When running large datasets in GLUE Full and SuperGLUE Full (MNLI, QQP, QNLI, SST-2, BoolQ, MultiRC, and WiC) using BERT$_{\text{Large}}$, we use learning rates $c=2,4,6,8,10,14$ to accelerate the training. The batch sizes for GLUE 100, 1k, Full are 4, 8, 16, respectively. The batch size for SuperGLUE is 4 except that the BERT$_{\text{Large}}$ models use 8 in SuperGLUE 1k and Full. For BERT$_{\text{Base}}$, the warmup ratio is 0.1. For BERT$_{\text{Large}}$, the warmup ratio is 0.2 and the weight decay is $10^{-6}$. 

For each fine-tuning random seed, we randomly select a different subset in the settings where only 100 or 1k training samples are available. For the datasets with less than 500 training samples in SuperGLUE and SuperGLUE 1k (i.e., CB and COPA), we repeat the experiments 32 times to further stabilize the scores. For the pre-trained BERT baseline, we use 16 fine-tuning random seeds. To reduce the computational cost, we use two pretraining random seeds and four fine-tuning random seeds in our ablation study in \autoref{tb:NLU_ablation}.

Compared to other tasks, ReCoRD needs to be trained much longer than other tasks in SuperGLUE, so we only use one fine-tuning seed for each of the four pretrained models with different seeds. Our fine-tuning stops after 600k batches (BERT$_{\text{Base}}$) / 300k batches (BERT$_{\text{Large}}$) or consecutive 160k batches without a validation improvement (whichever comes first). 

To stabilize the performance of each model on ReCoRD, we use the first 40k validation samples to select the best fine-tuned model checkpoints. We set batch size as 8 and learning rate as $1 \cdot 10^{-5}$ for BERT$_{\text{Base}}$. For BERT$_{\text{Large}}$, we set batch size as 32 and learning rate as $2 \cdot 10^{-5}$.





\subsection{Ensemble Models}
\label{sec:ensembling_analysis_details}

\textbf{Ensemble on FT Seeds (K=1)} in \autoref{tb:NLU_ablation} is the same as  \textbf{Ensemble of Ours (K=1)} in \autoref{tb:NLU_ECE}. \textbf{Ensemble on FT Seeds (K=5)} in \autoref{tb:NLU_ablation} is the same as \textbf{ENS} in \autoref{tb:NLU_top20}. \textbf{Ensemble on Dropouts} in \autoref{tb:NLU_ablation} is the same as \textbf{Dropout} in \autoref{tb:NLU_top20}. All results are the average of four models that use four different pretrained models and the best learning rate among $c \cdot 10^{-5}$ ($c=1,2,3,4,5,7$) in the fine-tuning stage.

In \autoref{tb:NLU_ECE}, we compute the expected calibration error (ECE)~\citep{naeini2015obtaining} by 
\begin{equation}
\label{eq:ECE}
\sum_{j=1}^{10} \frac{|B_j|}{N} |\text{acc}(j) - \text{conf}(j)|,
\end{equation}
where $\text{acc}(j)$ is the model accuracy in the $j$th bin $B_j$, $N$ is the number of validation samples, and $\text{conf}(j)=\frac{1}{|B_j|} \sum_{x \in B_j} \text{max}_y P(y|x)$ is the average of the highest prediction probability $P(y|x)$ in the $j$th bin. We put the samples into 10 equal-size bins according to their highest prediction probability $\text{max}_y P(y|x)$.

In \autoref{tb:NLU_ECE}, we use Tesla M40 to measure the inference time of the models built on BERT$_{\text{Base}}$. We set batch size 16 and run 1000 batches to get the average inference time of one batch in every GLUE task. We repeat the experiments five times and report their average and standard error. For the ensemble model, we assume the time of averaging multiple prediction probabilities is negligible and directly multiply the inference time of \textbf{Ours (K=1)} by 5.





In \autoref{tb:NLU_top20}, we would like to see if CLS embeddings disagree with each other as other ensemble baselines did. In \textbf{Multi-CLS}, we compute the uncertainty of each sample $x$ as the average variance of prediction probability of each CLS embedding $\text{mean}_l \left( \text{var}_k P(y=l|x,k) \right)$ and estimate the prediction probability of the $k$th CLS embedding by
\begin{equation}
\label{eq:prob_cls}
P(y=l|x,k) = \frac{ \text{exp}\left(\vq_{l,k}^T L^{FT}_{O,k}(\vh^c_k(x,y_{gt}) ) \right) }{ \sum_i \text{exp}\left(\vq_{i,k}^T L^{FT}_{O,k}(\vh^c_k(x,y_{gt}) ) \right) },
\end{equation}
where $L^{FT}_{O,k}(\vh^c_k(x,y_{gt}) )$ is the CLS embedding of the input $x$ after fine-tuning, and $\vq_{i,k}=\frac{1}{N_i}\sum_{y_{gt}=i} L^{FT}_{O,k}( \vh^c_k(x,y_{gt}) )$ is the $i$th class embedding for the $k$th CLS embedding, which is computed by averaging the $k$th CLS embeddings of the input $x$ with the $i$th class label.

In \autoref{tb:NLU_top20}, the two ensemble models for \textbf{ENS vs ENS} use the same set of 5 fine-tuning seeds and the two \textbf{Ours (K=5, $\mathbf{\lambda=0.1}$)} pretrained with different random seeds. Both uncertainty estimation models for \textbf{Multi-CLS vs ENS}, \textbf{Dropout vs ENS}, and \textbf{Least vs ENS} are based on the same pretrained \textbf{Ours (K=5, $\mathbf{\lambda=0.1}$)} model.








\section{Visualization of CLS embeddings}
\label{sec:emb_vis}

Table \ref{CoLA_k5_exmpales}--\ref{STSB_K1_exmpales} compare the CLS embeddings of \textbf{Ours (K=1)} and \textbf{Ours (K=5, $\mathbf{\lambda=0.1}$)} after fine-tuning to illustrate how different CLS embeddings capture distinct aspects of an input sentence in solving a task. For each task, we select one sample (a sentence or a sentence pair) from the validation set whose CLS embeddings disagree with each other. 

For each selected sample, we visualize its nearest-neighboring sentences in the validation set with respect to each CLS embedding. The nearest neighbors for the $k$th CLS embedding are determined by the cosine similarity between the respective $k$th CLS embedding of the input sentence and other sentences. Beside each sentence or sentence pair, we show their ground truth label and the model's prediction.

In \textbf{Ours (K=5, $\mathbf{\lambda=0.1}$)}, two representative sentences are selected from the top-three nearest neighbors for each CLS, and each CLS is manually annotated with terms that summarize those aspects that are shared by the neighbors and relate to the query sentence. For comparison, accompanying tables show the top-ten nearest neighbors for \textbf{Ours (K=1)}.

In almost all the classification tasks, we observe that CLS 3 and CLS 5 vote for the same class (i.e., their embeddings are close to the neighbors with the same class prediction). On the other hand, CLS 1 and CLS 4 vote for another class in these examples where the CLS embeddings disagree. The observation suggests that the similarity of CLS embeddings after the pretraining stage correlates with their similarity after the fine-tuning. 

\begin{table*}
\centering
\small
\renewcommand{\arraystretch}{1.2}
\setlength\tabcolsep{2pt}
\begin{tabular}{M{0.1\textwidth} M{0.1\textwidth} m{0.65\textwidth} M{0.1\textwidth}}
\hline
\multicolumn{4}{l}{\textbf{Task}: CoLA} \\
\hline
\textbf{Prediction} & \textbf{Label} & \multicolumn{1}{c}{\textbf{Sentence}} & \multicolumn{1}{c}{\textbf{Summary}} \\
\hline

\multicolumn{4}{l}{\mycctwenty Query:} \\
{\centering{un- \par acceptable}} & {\centering{un- \par acceptable}} & \multicolumn{2}{l}{{I lent the book partway to Tony.}} \\

\multicolumn{4}{l}{\mycctwenty CLS-space Neighbors (\textbf{K}=\textbf{5}):} \\
 \multicolumn{4}{l}{\mycc CLS 1} \\
 {\centering{acceptable}} & {\centering{acceptable}} &  {I gave it to Pete to take to the fair.} & \multirow{2}{*}[-2mm]{{ \emph{Gave to} }}\\\cline{1-3}

{\centering{acceptable}} & {\centering{un- \par acceptable}} & {Sue gave to Bill a book.}  & \\

 \multicolumn{4}{l}{\mycc CLS 2} \\
 {\centering{un- \par acceptable}} & {\centering{un- \par acceptable}} & {We wanted to invite someone, but we couldn't decide who to.} &\multirow{2}{*}[-2mm]{\makecell{ Incorrect or \\ extra word }}\\\cline{1-3}
  
 {\centering{un- \par acceptable}} & {\centering{un- \par acceptable}}  & {Jessica crammed boxes at the truck.}  & \\

 \multicolumn{4}{l}{\mycc CLS 3} \\
 {\centering{un- \par acceptable}} & {\centering{un- \par acceptable}} & {We wanted to invite someone, but we couldn't decide who to.}  &
 \makecell{ Incorrect or \\ extra word } \\\cline{1-3}
 
 {\centering{un- \par acceptable}} & {\centering{un- \par acceptable}} & {I hit that you knew the answer.}  & First person \\

 \multicolumn{4}{l}{\mycc CLS 4} \\
 {\centering{acceptable}} & {\centering{acceptable}} & {The paper was written up by John.}  &\multirow{2}{*}[-0mm]{{ Writing }}\\\cline{1-3}

 {\centering{acceptable}} & {\centering{acceptable}} & {John owns the book.}  & \\

 \multicolumn{4}{l}{\mycc CLS 5} \\
 {\centering{un- \par acceptable}} & {\centering{un- \par acceptable}} & {Chris was handed Sandy a note by Pat.}  &
  \multirow{2}{*}[-0mm]{{ \makecell{Extra word \\ Writing \\ Giving}}} \\\cline{1-3}
 
{\centering{un- \par acceptable}} & {\centering{un- \par acceptable}} & {What Mary did Bill was give a book.}  & \\

\hline
\end{tabular}
\caption{Visualization of \textbf{Ours (K=5, $\mathbf{\lambda=0.1}$)} using a sample in CoLA. The neighbors from CLS 2, 3, and 5 are unacceptable sentences that often contain extra words, as in the query. The neighbors from CLS 1 and 2 are semantically related to the query.
} 
\label{CoLA_k5_exmpales}
\end{table*}

\begin{table*}
\centering
\small
\renewcommand{\arraystretch}{1.2}
\setlength\tabcolsep{2pt}
\begin{tabular}{M{0.1\textwidth} M{0.1\textwidth} m{0.75\textwidth}}
\hline
\multicolumn{3}{l}{\textbf{Task}: CoLA} \\
\hline
\textbf{Prediction} & \textbf{Label} & \multicolumn{1}{c}{\textbf{Sentence}} \\
\hline

\multicolumn{3}{l}{\mycctwenty Query:} \\
un- \par acceptable & un- \par acceptable & I lent the book partway to Tony. \\

\multicolumn{3}{l}{\mycctwenty CLS-space Neighbors (\textbf{K}=\textbf{1}):} \\
un- \par acceptable & un- \par acceptable & I presented John with it dead. \\\cline{1-3}

un- \par acceptable & acceptable & Nora sent the book. \\\cline{1-3}

un- \par acceptable & un- \par acceptable & There seemed to be intelligent. \\\cline{1-3}

un- \par acceptable & un- \par acceptable & The book what inspired them was very long. \\\cline{1-3}

un- \par acceptable & un- \par acceptable & The book was by John written. \\\cline{1-3}

un- \par acceptable & acceptable & I met the man who grows peaches. \\\cline{1-3}

un- \par acceptable & acceptable & We persuaded Mary to leave and Sue to stay. \\\cline{1-3}

un- \par acceptable & un- \par acceptable & I hit that you knew the answer. \\\cline{1-3}

un- \par acceptable & un- \par acceptable & We think that Leslie likes ourselves. \\\cline{1-3}

un- \par acceptable & acceptable & This flyer and that flyer differ. \\
\hline
\end{tabular}
\caption{
Visualization of \textbf{Ours (K=1)} using the sample in CoLA.
} 
\label{CoLA_k1_exmpales}
\end{table*}


\begin{table*}
\centering
\small
\renewcommand{\arraystretch}{1.2}
\setlength\tabcolsep{2pt}
\begin{tabular}{M{0.1\textwidth} M{0.1\textwidth} m{0.65\textwidth} M{0.1\textwidth}}
\hline
\multicolumn{4}{l}{\textbf{Task}: SST-2}\\
\hline
\textbf{Prediction} & \textbf{Label} & \multicolumn{1}{c}{\textbf{Sentence}} & \multicolumn{1}{c}{\textbf{Summary}} \\
\hline

\multicolumn{4}{l}{\mycctwenty Query:} \\
  {\centering{negative}} &  {\centering{negative}}  &   {{An occasionally funny, but overall limp, fish-out-of-water story.}}
    &\\

   \multicolumn{4}{l}{\mycctwenty CLS-space Neighbors (\textbf{K}=\textbf{5}):} \\
 \multicolumn{4}{l}{\mycc CLS 1} \\
  \multirow{3}{*}{\centering{negative}} &  \multirow{3}{*}{\centering{positive}} &  
  \multirow{3}{*}{\makecell[l]{Based on a devilishly witty script by Heather McGowan and Niels Mueller, the \\ film gets great laughs, but never at the expense of its characters}}
 & \multirow{5}{*}[-1mm]{{\makecell{ \emph{Fun} or \emph{funny} \\ \\ Commas \\ \\ \emph{Script}}  }}\\\\\\\cline{1-3}
 \multirow{2}{*}{\centering{negative}} &  \multirow{2}{*}{\centering{positive}} & 
 \multirow{2}{*}{{McConaughey's fun to watch, the dragons are okay, not much fire in the script.}}
& \\ \\

 \multicolumn{4}{l}{\mycc CLS 2} \\
  \multirow{3}{*}{\centering{negative}} & \multirow{3}{*}{\centering{negative}} & 
  \multirow{3}{*}{\makecell[l]{Visually rather stunning, but ultimately a handsome-looking bore, the true \\ creativity would have been to hide Treasure Planet entirely and completely \\ reimagine it.}}
  &\multirow{5}{*}[0mm]{{\makecell{ \emph{Stunning} or \\ \emph{thrilling} but \\ negative \\ \\ Sci-fi }}}\\ \\ \\\cline{1-3}

  
  
  \multirow{2}{*}{\makecell{\centering{negative}}} & \multirow{2}{*}{\makecell{\centering{negative}}} & \multirow{2}{*}{\makecell
  {{If looking for a thrilling sci-fi cinematic ride, don't settle for this imposter.}}}
  &\\ \\


 \multicolumn{4}{l}{\mycc CLS 3} \\
 \multirow{2}{*}{\makecell{{\centering{positive}}}} & \multirow{2}{*}{\makecell{{\centering{positive}}}} & \multirow{2}{*}{\makecell{
 {Funny but perilously slight.}}} 
 &\multirow{3}{*}[-1mm]{\makecell{Positive \\ overall \\ but \\ qualified }}\\\\\cline{1-3}

 
 {\centering{positive}} & {\centering{positive}} &
 {{A movie that successfully crushes a best selling novel into a timeframe that mandates that you avoid the Godzilla sized soda.}}
 &\\

 \multicolumn{4}{l}{\mycc CLS 4} \\
  \multirow{3}{*}{\centering{negative}} & \multirow{3}{*}{\centering{negative}} & \multirow{3}{*}{\makecell[l]{Passable entertainment, but it's the kind of motion picture that won't make much of \\ a splash when it's released, and will not be remembered long afterwards.}} & \multirow{6}{*}[0mm]{\makecell{ Positive \\ statement, \\ \emph{but} \\ negative \\ \\  Liquid }}\\ \\ \\\cline{1-3}

 \multirow{3}{*}{\centering{negative}} & \multirow{3}{*}{\centering{negative}} & \multirow{3}{*}{\makecell[l]{It showcases Carvey's talent for voices, but not nearly enough and not without \\ taxing every drop of one's patience to get to the good stuff.}} & \\ \\ \\

 \multicolumn{4}{l}{\mycc CLS 5} \\
 {\centering{positive}} & {\centering{positive}} & {{The terrific and bewilderingly underrated Campbell Scott gives a star performance that is nothing short of mesmerizing.}} &
 \multirow{4}{*}[-0mm]{\makecell{ \emph{Mesmeriz-} \\ \emph{ing} or \\ \emph{intense} \\ and \\ positive }} \\\cline{1-3}
 
 \multirow{3}{*}{\centering{positive}} & \multirow{3}{*}{\centering{positive}} & \multirow{3}{*}{\makecell[l]{... an otherwise intense, twist-and-turn thriller that certainly shouldn't hurt talented \\ young Gaghan's resume.}}  & \\ \\ \\

\hline
\end{tabular}
\caption{
Visualization of \textbf{Ours (K=5, $\mathbf{\lambda=0.1}$)} using a sample in SST-2. The neighbors from CLS 2 and 4 share the same "postive, but negative" template as in the query.  Like the query, CLS 1, 3, and 5 capture the positive aspects. Some CLSs also capture the semantic aspects of the query such as \emph{script}, \emph{sci-fi}, or \emph{liquid}.
} 
\label{SST2_K5_exmpales}
\end{table*}


\begin{table*}
\centering
\small
\renewcommand{\arraystretch}{1.2}
\setlength\tabcolsep{2pt}
\begin{tabular}{M{0.1\textwidth} M{0.1\textwidth} m{0.75\textwidth}}
\hline
\multicolumn{3}{l}{\textbf{Task}: SST-2}\\\hline
\textbf{Prediction} & \textbf{Label} & \multicolumn{1}{c}{\textbf{Sentence}}\\\hline
\multicolumn{3}{l}{\mycctwenty Query:} \\
{\centering{positive}} & {\centering{negative}}  &  {{An occasionally funny, but overall limp, fish-out-of-water story.}} \\

\multicolumn{3}{l}{\mycctwenty CLS-space Neighbors (\textbf{K}=\textbf{1}):} \\
{\centering{positive}} & {\centering{positive}} & {{In a way, the film feels like a breath of fresh air, but only to those that allow it in.}} \\\hline

{\centering{positive}} & {\centering{positive}} & {{A painfully funny ode to bad behavior.}} \\\hline

{\centering{positive}} & {\centering{positive}} & {{Two hours fly by -- opera's a pleasure when you don't have to endure intermissions -- and even a novice to the form comes away exhilarated.}} \\\hline

{\centering{positive}} & {\centering{positive}} & {{Huston nails both the glad-handing and the choking sense of hollow despair.}} \\\hline

{\centering{positive}} & {\centering{positive}} & {{The movie's relatively simple plot and uncomplicated morality play well with the affable cast.}} \\\hline

{\centering{positive}} & {\centering{positive}} & {{So much facile technique, such cute ideas, so little movie.}} \\\hline

{\centering{positive}} & {\centering{positive}} & {{A psychological thriller with a genuinely spooky premise and an above-average cast, actor Bill Paxton's directing debut is a creepy slice of gothic rural Americana.}} \\\hline

{\centering{positive}} & {\centering{positive}} & {{The primitive force of this film seems to bubble up from the vast collective memory of the combatants.}} \\\hline

{\centering{positive}} & {\centering{positive}} & {{The continued good chemistry between Carmen and Juni is what keeps this slightly disappointing sequel going, with enough amusing banter -- blessedly curse-free -- to keep both kids and parents entertained.}} \\\hline

{\centering{positive}} & {\centering{positive}} & {{This flick is about as cool and crowd-pleasing as a documentary can get.}} \\\hline

\end{tabular}
\caption{
Visualization of \textbf{Ours (K=1)} using the sample in SST-2.
}
\label{SST2_K1_exmpales}
\end{table*}

\begin{table*}
\centering
\small
\renewcommand{\arraystretch}{1.2}
\setlength\tabcolsep{2pt}
\begin{tabular}{M{0.1\textwidth} M{0.1\textwidth} m{0.65\textwidth} M{0.1\textwidth}}
\hline
\multicolumn{4}{l}{\textbf{Task}: MRPC} \\
\hline
\textbf{Prediction} & \textbf{Label} & \multicolumn{1}{c}{\textbf{Sentence Pair}} & \multicolumn{1}{c}{\textbf{Summary}} \\
\hline

\multicolumn{4}{l}{\mycctwenty Query:} \\
 equivalent & equivalent &
  \textbf{S1}: A man arrested for allegedly threatening to shoot and kill a city councilman from Queens was ordered held on \$100,000 bail during an early morning court appearance Saturday. \newline \textbf{S2}: The Queens man arrested for allegedly threatening to shoot City Councilman Hiram Monserrate was held on \$100,000 bail Saturday, a spokesman for the Queens district attorney said. & \\

  \multicolumn{4}{l}{\mycctwenty CLS-space Neighbors (\textbf{K}=\textbf{5}):} \\
 \multicolumn{4}{l}{\mycc CLS 1} \\
 equivalent & equivalent & \textbf{S1}: Myanmar's pro-democracy leader Aung San Suu Kyi will return home late Friday but will remain in detention after recovering from surgery at a Yangon hospital, her personal physician said.\newline \textbf{S2}: Myanmar's pro-democracy leader Aung San Suu Kyi will be kept under house arrest following her release from a hospital where she underwent surgery, her personal physician said Friday. & \multirow{2}{*}[-1mm]{\makecell{Comments \\ \\ Politics \\ \\ Justice }} \\\cline{1-3}

not \par equivalent & not \par equivalent & \textbf{S1}: Bob Richter, a spokesman for House Speaker Tom Craddick, had no comment about the ruling. \newline \textbf{S2}: Bob Richter, spokesman for Craddick, R-Midland, said the speaker had not seen the ruling and could not comment. & \\

 \multicolumn{4}{l}{\mycc CLS 2} \\
  equivalent & equivalent & \textbf{S1}: They were being held Sunday in the Camden County Jail on \$100,000 bail. \newline \textbf{S2}: They remained in Camden County Jail on Sunday on \$100,000 bail.&\multirow{2}{*}[-3mm]{\makecell{ Thousands \\ \\ Crime or \\ threat }}\\\cline{1-3}

  equivalent & equivalent & \textbf{S1}: "More than 70,000 men and women from bases in Southern California were deployed in Iraq. \newline \textbf{S2}: In all, more than 70,000 troops based in Southern California were deployed to Iraq. &\\

 \multicolumn{4}{l}{\mycc CLS 3} \\
 equivalent & equivalent & \textbf{S1}: Robert Walsh, 40, remained in critical but stable condition Friday at Staten Island University Hospital's north campus. \newline \textbf{S2}: Walsh, also 40, was in critical but stable condition at Staten Island University Hospital last night. &\multirow{2}{*}[-6mm]{\makecell{ Time }}\\\cline{1-3}

 equivalent & equivalent & \textbf{S1}: Blair's Foreign Secretary Jack Straw was to take his place on Monday to give a statement to parliament on the European Union. \newline \textbf{S2}: Blair's office said his Foreign Secretary Jack Straw would take his place on Monday to give a statement to parliament on the EU meeting the prime minister attended last week. &\\

 \multicolumn{4}{l}{\mycc CLS 4} \\
 not \par equivalent & not \par equivalent & \textbf{S1}: Franklin County Judge-Executive Teresa Barton said a firefighter was struck by lightning and was taken to the Frankfort Regional Medical Center. \newline \textbf{S2}: A county firefighter, was struck by lightning and was in stable condition at Frankfort Regional Medical Center. & \multirow{2}{*}[-5mm]{\makecell{ Comments \\ \\  Medical or \\  justice }}\\\cline{1-3}


 equivalent & equivalent & \textbf{S1}: Myanmar's pro-democracy leader Aung San Suu Kyi will return home late Friday but will remain in detention after recovering from surgery at a Yangon hospital, her personal physician said.\newline \textbf{S2}: Myanmar's pro-democracy leader Aung San Suu Kyi will be kept under house arrest following her release from a hospital where she underwent surgery, her personal physician said Friday. & \\

 \multicolumn{4}{l}{\mycc CLS 5} \\
 equivalent & equivalent & \textbf{S1}: Unable to find a home for him, a judge told mental health authorities they needed to find supervised housing and treatment for DeVries somewhere in California. \newline \textbf{S2}: The judge had told the state Department of Mental Health to find supervised housing and treatment for DeVries somewhere in California. &
 \multirow{2}{*}[-6mm]{\makecell{ Court's \\ ruling }} \\\cline{1-3}

equivalent & equivalent & \textbf{S1}: A former employee of a local power company pleaded guilty Wednesday to setting off a bomb that knocked out a power substation during the Winter Olympics last year. \newline \textbf{S2}: A former Utah Power meter reader pleaded guilty Wednesday to bombing a power substation during the 2002 Winter Olympics. & \\

\hline
\end{tabular}
\caption{Visualization of \textbf{Ours (K=5, $\mathbf{\lambda=0.1}$)} using a sample in MRPC. The neighbors from CLS 2, 3, and 5 focus on different aspects of the query. The neighbors from CLS 1 and 4 are someone's comments as in the query and might not be equivalent. Several CLSs are also related to justice.
} 
\label{MRPC_k5_exmpales}
\end{table*}


\begin{table*}
\centering
\small
\renewcommand{\arraystretch}{1.2}
\setlength\tabcolsep{2pt}
\begin{tabular}{M{0.1\textwidth} M{0.1\textwidth} m{0.75\textwidth}}
\hline
\multicolumn{3}{l}{\textbf{Task}: MRPC}\\
\hline
\textbf{Prediction} & \textbf{Label} & \multicolumn{1}{c}{\textbf{Sentence Pair}} \\
\hline

\multicolumn{3}{l}{\mycctwenty Query:} \\
equivalent & equivalent &
  \textbf{S1}: A man arrested for allegedly threatening to shoot and kill a city councilman from Queens was ordered held on \$100,000 bail during an early morning court appearance Saturday. \newline \textbf{S2}: The Queens man arrested for allegedly threatening to shoot City Councilman Hiram Monserrate was held on \$100,000 bail Saturday, a spokesman for the Queens district attorney said. \\

\multicolumn{3}{l}{\mycctwenty CLS-space Neighbors (\textbf{K}=\textbf{1}):} \\
equivalent & equivalent & \textbf{S1}: The Justice Department Aug. 19 gave pre-clearance for the Oct. 7 date for the election to recall Gov. Gray Davis, saying it would not affect minority voting rights. \newline \textbf{S2}: The Justice Department on Aug. 19 sanctioned the Oct. 7 date for recall election, saying it would not affect voting rights. \\\hline

equivalent & equivalent & \textbf{S1}: The worm attacks Windows computers via a hole in the operating system, an issue Microsoft on July 16 had warned about. \newline \textbf{S2}: The worm attacks Windows computers via a hole in the operating system, which Microsoft warned of 16 July. \\\hline

equivalent & equivalent & \textbf{S1}: O'Brien was charged with leaving the scene of a fatal accident, a felony. \newline \textbf{S2}: Bishop Thomas O'Brien, 67, was booked on a charge of leaving the scene of a fatal accident. \\\hline

equivalent & equivalent & \textbf{S1}: "There is no conscious policy of the United States, I can assure you of this, to move the dollar at all," he said. \newline \textbf{S2}: He also said there is no conscious policy by the United States to move the value of the dollar. \\\hline

equivalent & equivalent & \textbf{S1}: The AFL-CIO is waiting until October to decide if it will endorse a candidate. \newline \textbf{S2}: The AFL-CIO announced Wednesday that it will decide in October whether to endorse a candidate before the primaries. \\\hline

equivalent & equivalent & \textbf{S1}: Speaking for the first time yesterday, Brigitte's maternal aunt said his family was unaware he had was in prison or that he had remarried. \newline \textbf{S2}: Brigitte's maternal aunt said his family was unaware he had been sent to prison, or that he had remarried in Sydney. \\\hline

equivalent & not \par equivalent & \textbf{S1}: Rosenthal is hereby sentenced to custody of the Federal Bureau of prisons for one day with credit for time served," Breyer said to tumultuous cheers in the courtroom. \newline \textbf{S2}: "Rosenthal is hereby sentenced to custody of the Federal Bureau of Prisons for one day with credit for time served." \\\hline

equivalent & equivalent & \textbf{S1}: Police say CIBA was involved in the importation of qat, a narcotic substance legal in Britain but banned in the United States. \newline \textbf{S2}: Mr McKinlay said that CIBA was involved in the importation of qat, a narcotic substance legal in Britain but banned in the US. \\\hline

equivalent & equivalent & \textbf{S1}: Judge Craig Doran said it wasn't his role to determine if Hovan was "an evil man" but maintained that "he has committed an evil act." \newline \textbf{S2}: Judge Craig Doran said he couldn't determine if Hovan was "an evil man" but said he "has committed an evil act." \\\hline

equivalent & equivalent & \textbf{S1}: But MTA officials appropriated the money to the 2003 and 2004 budgets without notifying riders or even the MTA board members considering the 50-cent hike, Hevesi found. \newline \textbf{S2}: MTA officials appropriated the surplus money to later years' budgets without notifying riders or the MTA board members when the 50-cent hike was being considered, he said. \\\hline
\end{tabular}
\caption{
Visualization of \textbf{Ours (K=1)} using the sample in MRPC.
}
\label{MRPC_k1_exmpales}
\end{table*}


\begin{table*}
\centering
\small
\renewcommand{\arraystretch}{1.2}
\setlength\tabcolsep{2pt}
\begin{tabular}{M{0.1\textwidth} M{0.1\textwidth} m{0.65\textwidth} M{0.1\textwidth}}
\hline
\multicolumn{4}{l}{\textbf{Task}: MNLI}\\
\hline
\textbf{Prediction} & \textbf{Label} & \multicolumn{1}{c}{\textbf{Sentence Pair}} & \multicolumn{1}{c}{\textbf{Summary}} \\
\hline

\multicolumn{4}{l}{\mycctwenty Query:} \\
contradic-\par tion & contradic-\par tion & \textbf{S1}: There is very little left of old Ocho  the scant remains of Ocho Rios Fort are probably the oldest and now lie in an industrial area, almost forgotten as the tide of progress has swept over the town. \newline \textbf{S2}: There is nothing left of the Ocho Rios Fort. & \\

  \multicolumn{4}{l}{\mycctwenty CLS-space Neighbors (\textbf{K}=\textbf{5}):} \\
 \multicolumn{4}{l}{\mycc CLS 1} \\ 
 neutral & contradic-\par tion & \textbf{S1}: After the purge of foreigners, only a few stayed on, strictly confined to Dejima Island in Nagasaki Bay. \newline \textbf{S2}: A few foreigners were left free after the purge on foreigners. & \multirow{2}{*}[-4mm]{\makecell{ Size or  \\ quantity }}\\\cline{1-3}

 neutral & neutral & \textbf{S1}: 'Publicity.' Lincoln removed his great hat, making a small show of dusting it off. \newline \textbf{S2}: Lincoln took his hat off. & \\

 \multicolumn{4}{l}{\mycc CLS 2} \\ 
   neutral & neutral & \textbf{S1}: There is no tradition of clothes criticism that includes serious analysis, or even of costume criticism among theater, ballet, and opera critics, who do have an august writerly heritage. \newline \textbf{S2}: Clothes criticism is not serious. &\multirow{2}{*}[2mm]{\makecell{ Historical \\ places \\ or \\ heritage \\  \\ Negation }}\\\cline{1-3}
  
   neutral & neutral & \textbf{S1}: All of the islands are now officially and proudly part of France, not colonies as they were for some three centuries. \newline \textbf{S2}: The islands are part of France now instead of just colonies. &\\

 \multicolumn{4}{l}{\mycc CLS 3} \\ 
  contradic-\par tion & neutral & \textbf{S1}: And yet, we still lack a set of global accounting and reporting standards that reflects the globalization of economies, enterprises, and markets. \newline \textbf{S2}: The globalization of economies is not reflected in global accounting standards. &\multirow{2}{*}[0mm]{\makecell{ Industry \\ \\ Region \\ \\ Negation }}\\\cline{1-3}
 
  contradic-\par tion & contradic-\par tion & \textbf{S1}: The technology used to capture and evaluate information in response to the RFP permits LSC to compile and assess key information about the delivery system at the program, state, regional, and national level. \newline \textbf{S2}: There is no way for the LSC to compile information about delivery systems. &\\

 \multicolumn{4}{l}{\mycc CLS 4} \\ 
  \multirow{3}{*}{\centering{neutral}} & \multirow{3}{*}{\centering{neutral}} & \multirow{3}{*}{\makecell[l]{\textbf{S1}: Scotland became little more than an English county. \\ \textbf{S2}: Scotland was hardly better than an English county.}} &\multirow{6}{*}[-0mm]{\makecell{ Historical \\ places \\ \\ Minimiza-\\ tion or \\ negation }}\\ \\ \\ \cline{1-3}

  \multirow{3}{*}{\centering{neutral}} & \multirow{3}{*}{\centering{neutral}} & \multirow{3}{*}{\makecell[l]{\textbf{S1}: Just as in ancient times, without the River Nile, Egypt could not exist. \\ \textbf{S2}: Without the Nile River, Egypt could not exist.}} & \\ \\ \\

 \multicolumn{4}{l}{\mycc CLS 5} \\
 {\centering{contradic-\par tion}} & {\centering{neutral}} & {{\textbf{S1}: Beside the fortress lies an 18th-century caravanserai, or inn, which has been converted into a hotel, and now hosts regular folklore evenings of Turkish dance and music. \newline \textbf{S2}: The 18th century caravanserai is now a hotel. }} &
 \multirow{2}{*}[0mm]{\makecell{ Buildings \\ or \\ \emph{properties} \\ \\ Contrast}} \\\cline{1-3}
 
 {\centering{contradic-\par tion}} & {\centering{contradic-\par tion}} & {{\textbf{S1}: Diamonds are graded from D to X, with only D, E, and F considered good, D being colorless or river white, J slightly tinted, Q light yellow, and S to X yellow. \newline \textbf{S2}: There is no difference between diamonds, all having the same properties.}} & \\

\hline
\end{tabular}

\caption{Visualization of \textbf{Ours (K=5, $\mathbf{\lambda=0.1}$)} using a sample in MNLI. Only one sentence in the neighbors of CLS 3 contains negation. Only the premise in the neighbors from CLS 5 makes a comparison. Both CLSs vote for the contradiction class. Several CLSs are related to buildings or historical places.
}
\label{MNLI_K5_exmpales}
\end{table*}

\begin{table*}
\centering
\small
\renewcommand{\arraystretch}{1.2}
\setlength\tabcolsep{2pt}
\begin{tabular}{M{0.1\textwidth} M{0.1\textwidth} m{0.75\textwidth}}
\hline
\multicolumn{3}{l}{\textbf{Task}: MNLI}\\
\hline
\textbf{Prediction} & \textbf{Label} & \multicolumn{1}{c}{\textbf{Sentence Pair}} \\
\hline

\multicolumn{3}{l}{\mycctwenty Query:} \\
contradic-\par tion & contradic-\par tion & \textbf{S1}: There is very little left of old Ocho  the scant remains of Ocho Rios Fort are probably the oldest and now lie in an industrial area, almost forgotten as the tide of progress has swept over the town. \newline \textbf{S2}: There is nothing left of the Ocho Rios Fort.
 \\
\multicolumn{3}{l}{\mycctwenty CLS-space Neighbors (\textbf{K}=\textbf{1}):} \\
contradic-\par tion & contradic-\par tion & \textbf{S1}: It was utterly mad. \newline \textbf{S2}: It was perfectly normal. \\\hline
contradic-\par tion & neutral & \textbf{S1}: Fixing current levels of damage would be impossible. \newline \textbf{S2}: Fixing the damage could never be done. \\\hline

contradic-\par tion & contradic-\par tion & \textbf{S1}: It was still night. \newline \textbf{S2}: The sun was blazing in the sky, darkness nowhere to be seen. \\\hline

contradic-\par tion & contradic-\par tion & \textbf{S1}: That's their signal \newline \textbf{S2}: That isn't their signal. \\\hline

contradic-\par tion & contradic-\par tion & \textbf{S1}: It is extremely dangerous to  Every trip to the store becomes a temptation. \newline \textbf{S2}: Even with every trip to the store, it never becomes a temptation. \\\hline

contradic-\par tion & contradic-\par tion & \textbf{S1}: The Revolutionaries couldn't be dissuaded from destroying most of the cathedral's statues, although 67 were saved (many of the originals are now housed in the Mus\'{e}e de l'Oeuvre Notre-Dame next door). \newline \textbf{S2}: All of the cathedrals statues were saved by the Revolutionaries. \\\hline

contradic-\par tion & contradic-\par tion & \textbf{S1}: It was deserved. \newline \textbf{S2}: It was not deserved at all \\\hline

contradic-\par tion & entailment & \textbf{S1}: And far, far away- lying still on the tracks- was the back of the train. \newline \textbf{S2}: The train wasn't moving but then it started up. \\\hline

contradic-\par tion & contradic-\par tion & \textbf{S1}: Even if you're the kind of traveler who likes to improvise and be adventurous, don't turn your nose up at the tourist offices. \newline \textbf{S2}: There's nothing worth seeing in the tourist offices. \\\hline

contradic-\par tion & contradic-\par tion & \textbf{S1}: Cybernetics had always been Derry's passion. \newline \textbf{S2}: Derry knew nothing of cybernetics. \\\hline

\end{tabular}
\caption{
Visualization of \textbf{Ours (K=1)} using the sample in MNLI.
}
\label{MNLI_K1_exmpales}
\end{table*}


\begin{table*}
\centering
\small
\renewcommand{\arraystretch}{1.2}
\setlength\tabcolsep{2pt}
\begin{tabular}{M{0.1\textwidth} M{0.1\textwidth} m{0.65\textwidth} M{0.1\textwidth}}
\hline
\multicolumn{4}{l}{\textbf{Task}: QNLI}\\
\hline
\textbf{Prediction} & \textbf{Label} & \multicolumn{1}{c}{\textbf{Sentence Pair}} & \multicolumn{1}{c}{\textbf{Summary}} \\
\hline

\multicolumn{4}{l}{\mycctwenty Query:} \\
entailment & entailment & \textbf{S1}: What factors negatively impacted Jacksonville following the war? \newline \textbf{S2}: Warfare and the long occupation left the city disrupted after the war. \\

\multicolumn{4}{l}{\mycctwenty CLS-space Neighbors (\textbf{K}=\textbf{5}):} \\
\multicolumn{4}{l}{\mycc CLS 1} \\
entailment & entailment & \textbf{S1}: When was the Russian policy "indigenization" defunded?\newline
\textbf{S2}: Never formally revoked, it stopped being implemented after 1932. & \multirow{2}{*}[-4mm]{{\centering{ Time}}}\\\cline{1-3}

entailment & entailment & \textbf{S1}: How did Luther describe his learning at the university? \newline \textbf{S2}: He was made to wake at four every morning for what has been described as "a day of rote learning and often wearying spiritual exercises." & \\

\multicolumn{4}{l}{\mycc CLS 2} \\
entailment & not \par entailment & \textbf{S1}: How did the 2001 IPCC report compare to reality for 2001-2006? \newline \textbf{S2}: The study compared IPCC 2001 projections on temperature and sea level change with observations. &\multirow{2}{*}[-4mm]{{\centering{Change }}}\\\cline{1-3}

entailment & entailment & \textbf{S1}: Who led the most rapid expansion of the Mongol empire? \newline \textbf{S2}: Under Genghis's successor Ogedei Khan the speed of expansion reached its peak. &\\

\multicolumn{4}{l}{\mycc CLS 3} \\
not \par entailment & not\par entailment & \textbf{S1}: During which period did Jacksonville become a popular destination for the rich? \newline \textbf{S2}: This highlighted the visibility of the state as a worthy place for tourism. &\multirow{2}{*}[-1mm]{\makecell{ \emph{Jacksonville} \\ \\ Duration of  \\ time}}\\\cline{1-3}

not \par entailment & not \par entailment & \textbf{S1}: What brought the downfall of Jacksonville filmmaking? \newline \textbf{S2}: Over the course of the decade, more than 30 silent film studios were established, earning Jacksonville the title of "winter film capital of the world". &\\

\multicolumn{4}{l}{\mycc CLS 4} \\
entailment & entailment & \textbf{S1}: How did the new king react to the Huguenots? \newline \textbf{S2}: Louis XIV gained the throne in 1643 and acted increasingly aggressively to force the Huguenots to convert. &\multirow{2}{*}[-2mm]{{\centering{ Change }}}\\\cline{1-3}

entailment & not \par entailment & \textbf{S1}: What did Luther begin to experience in 1536? \newline \textbf{S2}: In December 1544, he began to feel the effects of angina. & \\

\multicolumn{4}{l}{\mycc CLS 5} \\
not \par entailment & not \par entailment & \textbf{S1}: What brought the downfall of Jacksonville filmmaking? \newline \textbf{S2}: Over the course of the decade, more than 30 silent film studios were established, earning Jacksonville the title of "winter film capital of the world". &
\multirow{2}{*}[-4mm]{\makecell{ Negative \\ event}} \\\cline{1-3}
 
not \par entailment & not \par entailment & \textbf{S1}: What cycle AC current system did Tesla propose? \newline \textbf{S2}: He found the time there frustrating because of conflicts between him and the other Westinghouse engineers over how best to implement AC power. & \\

\hline
\end{tabular}
\caption{Visualization of \textbf{Ours (K=5, $\mathbf{\lambda=0.1}$)} using a sample in QNLI. The neighbors from CLS 1, 2, and 4 are about time or changes. The neighbors from CLS 3 and 5 are about \emph{Jacksonville} or negative events.}
\label{QNLI_K5_exmpales}
\end{table*}


\begin{table*}
\centering
\small
\renewcommand{\arraystretch}{1.2}
\setlength\tabcolsep{2pt}
\begin{tabular}{M{0.1\textwidth} M{0.1\textwidth} m{0.75\textwidth}}
\hline
\multicolumn{3}{l}{\textbf{Task}: QNLI} \\
\hline
\textbf{Prediction} & \textbf{Label} & \multicolumn{1}{c}{\textbf{Sentence Pair}} \\
\hline

\multicolumn{3}{l}{\mycctwenty Query:} \\
entailment & entailment & \textbf{S1}: What factors negatively impacted Jacksonville following the war? \newline \textbf{S2}: Warfare and the long occupation left the city disrupted after the war. \\

\multicolumn{3}{l}{\mycctwenty CLS-space Neighbors (\textbf{K}=\textbf{1}):} \\
entailment & entailment & \textbf{S1}: How many Africans were brought into the United States during the slave trade? \newline \textbf{S2}: Participation in the African slave trade and the subsequent treatment of its 12 to 15 million Africans is viewed by some to be a more modern extension of America's "internal colonialism". \\\hline

entailment & entailment & \textbf{S1}: Which country used to rule California? \newline \textbf{S2}: Though there is no official definition for the northern boundary of southern California, such a division has existed from the time when Mexico ruled California, and political disputes raged between the Californios of Monterey in the upper part and Los Angeles in the lower part of Alta California. \\\hline

entailment & entailment & \textbf{S1}: In what area of this British colony were Huguenot land grants? \newline \textbf{S2}: In 1700 several hundred French Huguenots migrated from England to the colony of Virginia, where the English Crown had promised them land grants in Lower Norfolk County. \\\hline

entailment & entailment & \textbf{S1}: Who was responsible for the new building projects in Jacksonville? \newline \textbf{S2}: Mayor W. Haydon Burns' Jacksonville Story resulted in the construction of a new city hall, civic auditorium, public library and other projects that created a dynamic sense of civic pride. \\\hline

entailment & entailment & \textbf{S1}: What did Tesla first receive after starting his company? \newline \textbf{S2}: The company installed electrical arc light based illumination systems designed by Tesla and also had designs for dynamo electric machine commutators, the first patents issued to Tesla in the US. \\\hline

entailment & entailment & \textbf{S1}: In what year did the university first see a drop in applications? \newline \textbf{S2}: In the early 1950s, student applications declined as a result of increasing crime and poverty in the Hyde Park neighborhood. \\\hline

entailment & entailment & \textbf{S1}: What was Fresno's population in 2010? \newline \textbf{S2}: The 2010 United States Census reported that Fresno had a population of 494,665. \\\hline

entailment & entailment & \textbf{S1}: What was the percentage of Black or African-Americans living in the city? \newline \textbf{S2}: The racial makeup of the city was 50.2\% White, 8.4\% Black or African American, 1.6\% Native American, 11.2\% Asian (about a third of which is Hmong), 0.1\% Pacific Islander, 23.4\% from other races, and 5.2\% from two or more races. \\\hline

entailment & entailment & \textbf{S1}: Where did Marin build first fort? \newline \textbf{S2}: He first constructed Fort Presque Isle (near present-day Erie, Pennsylvania) on Lake Erie's south shore. \\\hline

entailment & not \par entailment & \textbf{S1}: How old was Tesla when he became a US citizen? \newline \textbf{S2}: In the same year, he patented the Tesla coil. \\\hline
\end{tabular}
\caption{Visualization of \textbf{Ours (K=1)} using the sample in QNLI.
}
\label{QNLI_K1_exmpales}
\end{table*}


\begin{table*}
\centering
\small
\renewcommand{\arraystretch}{1.2}
\setlength\tabcolsep{2pt}
\begin{tabular}{M{0.1\textwidth} M{0.1\textwidth} m{0.65\textwidth} M{0.1\textwidth} }
\hline
\multicolumn{4}{l}{\textbf{Task}: STS-B}\\
\hline
\textbf{Prediction} & \textbf{Label} & \multicolumn{1}{c}{\textbf{Sentence Pair}} & \multicolumn{1}{c}{\textbf{Summary}} \\
\hline

\multicolumn{4}{l}{\mycctwenty Query:} \\
 2.009&  2.000 & \textbf{S1}:  Volkswagen skids into red in wake of pollution scandal \newline \textbf{S2}: Volkswagen's "gesture of goodwill" to diesel owners
    & \\
  \multicolumn{4}{l}{\mycctwenty CLS-space Neighbors (\textbf{K}=\textbf{5}):} \\

 \multicolumn{4}{l}{\mycc CLS 1} \\
 \multirow{3}{*}{2.633} &  \multirow{3}{*}{3.800} & \multirow{3}{*}{\makecell[l]{\textbf{S1}: Rosberg emulates father with Monaco win  \\ \textbf{S2}: FORMULA 1: Rosberg stays modest despite Monaco win }} & \multirow{4}{*}[-2mm]{\makecell{ Motor  \\ vehicles \\ \\ Racing }}\\\\\\\cline{1-3}

 2.754 & 3.000 & \textbf{S1}: A motorcross driver going by during a race \newline \textbf{S2}:  A race car driver performs in the race of his life. & \\

 \multicolumn{4}{l}{\mycc CLS 2} \\ 
  1.710 & 1.400 & {\textbf{S1}: A golden dog is running through the snow. \newline \textbf{S2}: A pack of sled dogs pulling a sled through a town. } &\multirow{2}{*}[-1mm]{\makecell{ Colors \\ \\ Action }}\\\cline{1-3}
  
  1.917 & 1.400 & {\textbf{S1}: The black and white dog is running on the grass.  \newline \textbf{S2}: A black and white dog swims in blue water.} &\\

 \multicolumn{4}{l}{\mycc CLS 3} \\
  2.952 & 2.600 & \textbf{S1}: Obama endorses same-sex marriage \newline \textbf{S2}: Obama's delicate dance on same-sex marriage &\multirow{2}{*}[-1mm]{\makecell{ Politics \\ and \\  economics }}\\\cline{1-3}
 
 2.071 & 1.000 & \textbf{S1}: Spanish jobless rate soars past 25 per cent \newline \textbf{S2}: US jobless rate seen rising, offering Obama no relief &\\

 \multicolumn{4}{l}{\mycc CLS 4} \\ 
  0.337 & 0.000 & \textbf{S1}: Presumably the decision of drivers to slow down in response to work zone signage is influenced by many factors. \newline \textbf{S2}: This short talk deals with issues of "cheating slightly" :Dan Ariely: Our buggy moral code . &\multirow{2}{*}[0mm]{\makecell{ Motor \\  vehicles \\ \\  Morality }}\\\cline{1-3}

  0.946 & 0.600 & \textbf{S1}: Saudi gas truck blast kills at least 22 \newline \textbf{S2}: Nigeria church blast kills at least 12 & \\

 \multicolumn{4}{l}{\mycc CLS 5} \\
 3.820 & 2.800 & \textbf{S1}: Stocks dipped lower Tuesday as investors opted to cash in profits from Monday's big rally despite a trio of reports suggesting modest improvement in the economy. \newline \textbf{S2}: Wall Street moved tentatively higher Tuesday as investors weighed a trio of reports showing modest economic improvement against an urge to cash in profits from Monday's big rally. &
 \multirow{2}{*}[-1mm]{\makecell{ Politics \\ and \\  economics  }} \\\cline{1-3}
 
2.952 & 2.600 & \textbf{S1}: Obama endorses same-sex marriage  \newline \textbf{S2}: Obama's delicate dance on same-sex marriage & \\

\hline
\end{tabular}
\caption{Visualization of \textbf{Ours (K=5, $\mathbf{\lambda=0.1}$)} using a sample in STS-B. The neighbors from CLS 1 and 4 are about motor vehicles. The neighbors from CLS 3 and 5 are about politics and economics.
} 
\label{STSB_K5_exmpales}
\end{table*}

\begin{table*}
\centering
\small
\renewcommand{\arraystretch}{1.2}
\setlength\tabcolsep{2pt}
\begin{tabular}{M{0.1\textwidth} M{0.1\textwidth} m{0.75\textwidth}}
\hline
\multicolumn{3}{l}{\textbf{Task}: STS-B} \\
\hline
\textbf{Prediction} & \textbf{Label} & \multicolumn{1}{c}{\textbf{Sentence Pair}} \\
\hline
\multicolumn{3}{l}{\mycctwenty Query:} \\
2.009 &  2.000 & \textbf{S1}:  Volkswagen skids into red in wake of pollution scandal \newline \textbf{S2}: Volkswagen's "gesture of goodwill" to diesel owners \\

\multicolumn{3}{l}{\mycctwenty CLS-space Neighbors (\textbf{K}=\textbf{1}):}\\
2.133 & 2.600 & \textbf{S1}: Large silver locomotive engine in a shed. \newline \textbf{S2}: The silver train is parked in a station. \\\hline

2.537 & 3.200 & \textbf{S1}: An AeroMexico jet taxing along a runway. \newline \textbf{S2}: a silver AreoMexico Jet Liner sitting on the tarmac. \\\hline

2.189 & 2.800 & \textbf{S1}: Two women holding checkered flags near an orange car. \newline \textbf{S2}: Two ladies in skimpy clothes pose next to an old fashioned car. \\\hline

1.963 & 2.400 & \textbf{S1}: Two dogs in the snow \newline \textbf{S2}: Two dogs play in the grass. \\\hline

2.041 & 1.600 & \textbf{S1}: Three dogs are playing in the white snow. \newline \textbf{S2}: Two dogs are playing in the grass. \\\hline

2.402 & 3.200 & \textbf{S1}: Once you open it up to toxins, the answer is clearly no, boiling is not enough. \newline \textbf{S2}: Boiling eliminates only a certain class of contaminants that can make you ill. \\\hline

1.990 & 1.400 & \textbf{S1}: A golden dog is running through the snow. \newline \textbf{S2}: A pack of sled dogs pulling a sled through a town. \\\hline

2.230 & 2.200 & \textbf{S1}: Man sitting on a bench drink from a mug surrounded by rugs. \newline \textbf{S2}: A man is sitting on one of two red benches and staring into a kiosk. \\\hline

2.426 & 2.000 & \textbf{S1}: If you can get over the "ick factor," you have an easily-applied source of organic nitrogen fertilizer close at hand. \newline \textbf{S2}: The NPK numbers on the fertilizer represents the percent, by weight, of Nitrogen, P2O5 and K2O, respectively. \\\hline

2.248 & 3.400 & \textbf{S1}: Try switching to rats; weanling rats if you need something smaller.  \newline \textbf{S2}: As mentioned in previous answers, rats and gerbils can be offered instead of mice or in a rotation with mice. \\\hline

\end{tabular}
\caption{Visualization of \textbf{Ours (K=1)} using the sample in STS-B.}
\label{STSB_K1_exmpales}
\end{table*}



\section{Diversity Measurement between two CLS embeddings}
\label{sec:diversity_metric}

We find that cosine similarities between the CLS embeddings are not a good measurement of their diversity. For different CLS $k$, if their hidden states $\vh^c_k$ are identical but their output linear layers  $L_{O,k}$ have different biases, the cosine similarity between CLS embeddings could be small but their diversity is also small.

Motivated by the visualization in \Cref{sec:emb_vis}, we found that the diversity between two CLS embeddings ($k_1$ and $k_2$) could be estimated by their similarity differences to their neighbors. If two CLS embeddings collapse, their dot products to their neighbors should perfectly correlated with each other and their resulting nearest neighbors would be the same. Thus, we estimate the diversity between CLS embeddings during pretraining by
\begin{align}
\label{eq:div_corr}
\text{Corr}\left([(\vc_{k_1,i}^{1-2})^T\vc_{k_1,j}^{3-4}]_{i,j}, [(\vc_{k_2,i}^{1-2})^T\vc_{k_2,j}^{3-4}]_{i,j}\right),
\end{align}
where $\vc_{k_1,i}^{1-2}$ is the $k_1$th CLS embedding of the $i$th sample for sentence 1 and 2 in the batch, $\vc_{k_1,j}^{3-4}$ is the $k_1$th CLS embedding of the $j$th sample for sentence 3 and 4 in the batch, $[(\vc_{k_1,i}^{1-2})^T\vc_{k_1,j}^{3-4}]_{i,j}$ is a sequence containing all the pairwise dot products of $k_1$th CLS embeddings in the batch, and Corr is the pearson correlation coefficient. Lower correlation means more diverse. 

We use the metric to test our diversification methods and 
detect the collapsing during pretraining. Without using the diversity tricks we developed (e.g., inserting linear layers into the transformer encoder), this metric would be greater than $0.99$ and the improvement would be greatly reduce in downstream applications (see our ablation study in \Cref{tb:NLU_ablation}). In contrast, our final best model reaches around $0.9$--$0.95$ in this metric. We found that if we use the fine-tuning re-parameterization trick during pretraining, we can have a lower correlation value (i.e., more diverse CLS embeddings), but the performance on GLUE is much worse. This indicates that there is an ideal diversity level for the consecutive sentence detection task during pretraining.


\section{Performance of Individual Tasks}
\label{sec:task_scores}

The GLUE tasks include CoLA \citep{warstadt2019neural},
SST-2 \citep{socher2013recursive}, MRPC \citep{dolan2005automatically} , QQP\footnote{\url{https://www.quora.com/profile/Ricky-Riche-2/First-Quora-Dataset-Release-Question-Pairs}}, STS-B \citep{cer2017semeval}, MNLI \citep{williams2018broad}, QNLI \citep{rajpurkar2016squad}, RTE \citep{bentivogli2009fifth}, and WNLI \citep{levesque2012winograd}. The SuperGLUE tasks include BoolQ \citep{clark2019boolq}, CB \citep{de2019commitmentbank}, COPA \citep{roemmele2011choice}, MultiRC \citep{khashabi2018looking}, ReCoRD \citep{zhang2018record}, RTE, WiC \citep{pilehvar2019wic}, and WSC \citep{levesque2012winograd}.

We report the individual task results of GLUE 100 and 1k in \autoref{tb:GLUE_all_scores}, the results of GLUE Full in \autoref{tb:GLUE_all_scores_full}, the results of SuperGLUE 100 and 1k in \autoref{tb:SuperGLUE_all_scores}, the results of SuperGLUE Full in \autoref{tb:SuperGLUE_all_scores_full}, the results of the top 20\% uncertain sample overlapping ratio in \autoref{tb:top_20_all}, and the results of ECE in \autoref{tb:ECE_all_scores}. In \autoref{tb:GLUE_all_scores_full}, we also compare the GLUE score of our \textbf{MTL} baseline with the scores reported in \citet{aroca2020losses}.

In GLUE 100 and SuperGLUE 100, multiple embeddings are almost always better. In GLUE 1k and Full, the improvement is smaller, so the baselines perform better in some individual tasks. 
We also observe that different downstream tasks might prefer different lambda. 

In \autoref{tb:ECE_all_scores}, we compute the p value using Chernoff bound: 
\begin{equation}
\label{eq:p_value}
P(X > (1+\delta)\mu ) < \left( \frac{ e^{\delta} }{ (1 + \delta)^{(1 + \delta)} }\right)^{\mu}, 
\end{equation}
where $\mu=(0.2)^2 4N$, $N$ is the number of samples in the validation set, $\delta = \frac{\sum_{i=1}^4 S_i}{\mu} -1$, and $S_i$ is the observed size of overlapping at the $i$th trial.


\begin{table*}[t!]
\scalebox{0.9}{
\begin{tabular}{|c|cccccccc|c|}
\multicolumn{10}{c}{GLUE 100 (BERT Base)} \\ \hline
  & CoLA &  SST &  MRPC &  STS-B &  QQP &  MNLI &  QNLI &  RTE &  Avg.\\ 
  & MCC  &  Acc &  F1   &  Spearman & F1 & Acc  &  Acc  &  Acc &    -  \\ \hline
Pretrained & \textbf{18.62} & 75.41 & 80.44 & 62.16 & 59.09 & 38.51 & 59.99 & 54.56 & 55.71\vspace{-0.2cm} \\
 & \VarCell{r}{1.96} & \VarCell{r}{1.95} & \VarCell{r}{0.57} & \VarCell{r}{3.68} & \VarCell{r}{0.94} & \VarCell{r}{0.69} & \VarCell{r}{1.33} & \VarCell{r}{0.79}\vspace{-0.1cm} & \VarCell{|r|}{0.62} \\
MTL & 9.90 & 70.67 & 81.64 & 78.88 & 59.74 & 43.50 & 73.54 & 57.49 & 59.29\vspace{-0.2cm} \\
 & \VarCell{r}{1.18} & \VarCell{r}{0.63} & \VarCell{r}{0.19} & \VarCell{r}{0.57} & \VarCell{r}{0.73} & \VarCell{r}{0.87} & \VarCell{r}{0.56} & \VarCell{r}{1.15}\vspace{-0.1cm} & \VarCell{|r|}{0.27} \\
Ours (K=1) & 11.24 & 70.24 & 80.97 & 78.10 & 58.94 & 40.99 & 68.69 & 55.18 & 57.84\vspace{-0.2cm} \\
 & \VarCell{r}{1.09} & \VarCell{r}{1.30} & \VarCell{r}{0.33} & \VarCell{r}{0.61} & \VarCell{r}{0.73} & \VarCell{r}{0.56} & \VarCell{r}{0.99} & \VarCell{r}{1.19}\vspace{-0.1cm} & \VarCell{|r|}{0.32} \\
Ours (K=5, $\lambda= 0$) & 17.44 & 74.31 & \textbf{81.98} & \textbf{79.53} & \textbf{61.98} & 44.47 & \textbf{75.94} & 58.44 & 61.54\vspace{-0.2cm} \\
 & \VarCell{r}{1.36} & \VarCell{r}{1.19} & \VarCell{r}{0.22} & \VarCell{r}{0.70} & \VarCell{r}{0.54} & \VarCell{r}{0.67} & \VarCell{r}{0.48} & \VarCell{r}{1.38}\vspace{-0.1cm} & \VarCell{|r|}{0.32} \\
Ours (K=5, $\lambda= 0.1$) & 17.61 & \textbf{75.49} & 81.68 & 79.25 & 61.70 & \textbf{46.09} & 75.12 & \textbf{59.17} & \textbf{61.80}\vspace{-0.2cm} \\
 & \VarCell{r}{1.75} & \VarCell{r}{0.96} & \VarCell{r}{0.14} & \VarCell{r}{0.66} & \VarCell{r}{0.63} & \VarCell{r}{0.84} & \VarCell{r}{0.56} & \VarCell{r}{1.48}\vspace{-0.1cm} & \VarCell{|r|}{0.35} \\
Ours (K=5, $\lambda= 0.5$) & 13.52 & 74.24 & 81.60 & 79.49 & 61.78 & 44.46 & 74.14 & 57.25 & 60.49\vspace{-0.2cm} \\
 & \VarCell{r}{1.94} & \VarCell{r}{0.95} & \VarCell{r}{0.15} & \VarCell{r}{0.51} & \VarCell{r}{0.49} & \VarCell{r}{0.62} & \VarCell{r}{0.66} & \VarCell{r}{1.34}\vspace{-0.1cm} & \VarCell{|r|}{0.35} \\
Ours (K=5, $\lambda= 1$) & 10.59 & 74.56 & 81.28 & 77.93 & 60.80 & 43.42 & 74.83 & 57.17 & 59.86\vspace{-0.2cm} \\
 & \VarCell{r}{1.68} & \VarCell{r}{0.89} & \VarCell{r}{0.16} & \VarCell{r}{0.67} & \VarCell{r}{0.82} & \VarCell{r}{0.95} & \VarCell{r}{0.70} & \VarCell{r}{1.20} & \VarCell{|r|}{0.34} \\
 \hline
\multicolumn{10}{c}{GLUE 100 (BERT Large)} \\ \hline
  & CoLA &  SST &  MRPC &  STS-B &  QQP &  MNLI &  QNLI &  RTE &  Avg.\\ \hline
MTL & 15.64 & 79.61 & 81.48 & 74.92 & 61.41 & 43.61 & 77.28 & 58.66 & 61.39\vspace{-0.2cm} \\
 & \VarCell{r}{1.60} & \VarCell{r}{1.63} & \VarCell{r}{0.19} & \VarCell{r}{0.90} & \VarCell{r}{0.47} & \VarCell{r}{0.86} & \VarCell{r}{0.53} & \VarCell{r}{1.24}\vspace{-0.1cm} & \VarCell{|r|}{0.37} \\
Ours (K=1) & 16.87 & 73.21 & 81.35 & 77.48 & 57.58 & 41.51 & 70.79 & 56.20 & 59.19\vspace{-0.2cm} \\
 & \VarCell{r}{1.99} & \VarCell{r}{1.57} & \VarCell{r}{0.18} & \VarCell{r}{0.63} & \VarCell{r}{0.75} & \VarCell{r}{0.87} & \VarCell{r}{1.90} & \VarCell{r}{0.57}\vspace{-0.1cm} & \VarCell{|r|}{0.43} \\
Ours (K=5, $\lambda= 0$) & \textbf{22.41} & 80.54 & 82.01 & 76.06 & 61.60 & 46.35 & 77.73 & 60.27 & 63.19\vspace{-0.2cm} \\
 & \VarCell{r}{1.97} & \VarCell{r}{2.11} & \VarCell{r}{0.23} & \VarCell{r}{1.97} & \VarCell{r}{0.96} & \VarCell{r}{0.86} & \VarCell{r}{0.64} & \VarCell{r}{1.23}\vspace{-0.1cm} & \VarCell{|r|}{0.49} \\
Ours (K=5, $\lambda= 0.1$) & 22.02 & \textbf{82.67} & 81.81 & \textbf{78.44} & \textbf{63.49} & 46.94 & 77.58 & \textbf{63.51} & \textbf{64.24}\vspace{-0.2cm} \\
 & \VarCell{r}{2.79} & \VarCell{r}{0.72} & \VarCell{r}{0.22} & \VarCell{r}{0.63} & \VarCell{r}{0.66} & \VarCell{r}{0.74} & \VarCell{r}{0.65} & \VarCell{r}{0.70}\vspace{-0.1cm} & \VarCell{|r|}{0.40} \\
Ours (K=5, $\lambda= 0.5$) & 18.59 & 80.47 & \textbf{82.02} & 77.16 & 61.18 & \textbf{47.04} & 77.43 & 61.91 & 63.02\vspace{-0.2cm} \\
 & \VarCell{r}{2.54} & \VarCell{r}{1.33} & \VarCell{r}{0.12} & \VarCell{r}{0.40} & \VarCell{r}{0.69} & \VarCell{r}{0.77} & \VarCell{r}{0.56} & \VarCell{r}{1.27}\vspace{-0.1cm} & \VarCell{|r|}{0.42} \\
Ours (K=5, $\lambda= 1$) & 15.76 & 79.98 & 81.83 & 76.73 & 62.27 & 45.27 & \textbf{77.99} & 59.24 & 62.07\vspace{-0.2cm} \\
 & \VarCell{r}{2.65} & \VarCell{r}{1.22} & \VarCell{r}{0.18} & \VarCell{r}{1.19} & \VarCell{r}{0.74} & \VarCell{r}{1.09} & \VarCell{r}{0.49} & \VarCell{r}{1.11} & \VarCell{|r|}{0.45} \\
 \hline
\multicolumn{10}{c}{GLUE 1k (BERT Base)} \\ \hline
  & CoLA &  SST &  MRPC &  STS-B &  QQP &  MNLI &  QNLI &  RTE &  Avg.\\ \hline
Pretrained & 42.71 & 87.08 & 86.98 & 85.93 & 70.01 & 58.05 & 77.97 & 64.34 & 71.67\vspace{-0.2cm} \\
 & \VarCell{r}{0.54} & \VarCell{r}{0.18} & \VarCell{r}{0.20} & \VarCell{r}{0.16} & \VarCell{r}{0.22} & \VarCell{r}{0.62} & \VarCell{r}{0.33} & \VarCell{r}{0.65}\vspace{-0.1cm} & \VarCell{|r|}{0.15} \\
MTL & 41.57 & 86.82 & 87.44 & 87.18 & 71.92 & 62.01 & 82.39 & 66.54 & 73.26\vspace{-0.2cm} \\
 & \VarCell{r}{0.68} & \VarCell{r}{0.16} & \VarCell{r}{0.24} & \VarCell{r}{0.15} & \VarCell{r}{0.20} & \VarCell{r}{0.26} & \VarCell{r}{0.26} & \VarCell{r}{0.53}\vspace{-0.1cm} & \VarCell{|r|}{0.13} \\
Ours (K=1) & 39.69 & 86.76 & 87.49 & 87.53 & 71.56 & 62.07 & 83.69 & 66.79 & 73.28\vspace{-0.2cm} \\
 & \VarCell{r}{0.63} & \VarCell{r}{0.21} & \VarCell{r}{0.31} & \VarCell{r}{0.10} & \VarCell{r}{0.28} & \VarCell{r}{0.19} & \VarCell{r}{0.23} & \VarCell{r}{0.58}\vspace{-0.1cm} & \VarCell{|r|}{0.13} \\
Ours (K=5, $\lambda= 0$) & 42.24 & 86.98 & 87.69 & 87.91 & \textbf{73.09} & \textbf{63.08} & \textbf{83.94} & 68.00 & \textbf{74.14}\vspace{-0.2cm} \\
 & \VarCell{r}{0.59} & \VarCell{r}{0.20} & \VarCell{r}{0.34} & \VarCell{r}{0.13} & \VarCell{r}{0.15} & \VarCell{r}{0.25} & \VarCell{r}{0.18} & \VarCell{r}{0.51}\vspace{-0.1cm} & \VarCell{|r|}{0.12} \\
Ours (K=5, $\lambda= 0.1$) & 42.60 & 86.90 & \textbf{87.76} & \textbf{88.05} & 72.81 & 62.63 & 83.68 & \textbf{68.10} & 74.10\vspace{-0.2cm} \\
 & \VarCell{r}{0.52} & \VarCell{r}{0.24} & \VarCell{r}{0.33} & \VarCell{r}{0.14} & \VarCell{r}{0.21} & \VarCell{r}{0.58} & \VarCell{r}{0.13} & \VarCell{r}{0.49}\vspace{-0.1cm} & \VarCell{|r|}{0.13} \\
Ours (K=5, $\lambda= 0.5$) & \textbf{42.75} & 86.78 & 87.55 & 87.88 & 72.56 & 62.71 & 83.66 & 68.00 & 74.02\vspace{-0.2cm} \\
 & \VarCell{r}{0.49} & \VarCell{r}{0.19} & \VarCell{r}{0.31} & \VarCell{r}{0.11} & \VarCell{r}{0.21} & \VarCell{r}{0.39} & \VarCell{r}{0.17} & \VarCell{r}{0.51}\vspace{-0.1cm} & \VarCell{|r|}{0.12} \\
Ours (K=5, $\lambda= 1$) & 40.08 & \textbf{87.36} & 87.54 & 87.74 & 72.83 & 62.79 & 83.53 & 67.98 & 73.75\vspace{-0.2cm} \\
 & \VarCell{r}{0.90} & \VarCell{r}{0.13} & \VarCell{r}{0.23} & \VarCell{r}{0.10} & \VarCell{r}{0.16} & \VarCell{r}{0.43} & \VarCell{r}{0.14} & \VarCell{r}{0.30} & \VarCell{|r|}{0.14} \\
 \hline
\multicolumn{10}{c}{GLUE 1k (BERT Large)} \\ \hline
  & CoLA &  SST &  MRPC &  STS-B &  QQP &  MNLI &  QNLI &  RTE &  Avg.\\ \hline
MTL & 49.10 & 89.84 & 87.53 & 87.85 & 73.04 & 62.70 & 84.74 & 67.52 & 75.30\vspace{-0.2cm} \\
 & \VarCell{r}{0.76} & \VarCell{r}{0.18} & \VarCell{r}{0.24} & \VarCell{r}{0.10} & \VarCell{r}{0.18} & \VarCell{r}{1.85} & \VarCell{r}{0.17} & \VarCell{r}{0.75}\vspace{-0.1cm} & \VarCell{|r|}{0.27} \\
Ours (K=1) & 46.89 & 89.54 & \textbf{88.41} & 87.61 & 72.58 & 64.51 & \textbf{85.20} & 67.61 & 75.35\vspace{-0.2cm} \\
 & \VarCell{r}{0.90} & \VarCell{r}{0.21} & \VarCell{r}{0.25} & \VarCell{r}{0.16} & \VarCell{r}{0.22} & \VarCell{r}{0.35} & \VarCell{r}{0.16} & \VarCell{r}{1.24}\vspace{-0.1cm} & \VarCell{|r|}{0.21} \\
Ours (K=5, $\lambda= 0$) & 49.76 & 89.93 & 87.38 & 87.91 & 72.65 & 63.50 & 85.00 & 69.66 & 75.73\vspace{-0.2cm} \\
 & \VarCell{r}{0.63} & \VarCell{r}{0.14} & \VarCell{r}{0.38} & \VarCell{r}{0.11} & \VarCell{r}{0.26} & \VarCell{r}{1.83} & \VarCell{r}{0.23} & \VarCell{r}{0.41}\vspace{-0.1cm} & \VarCell{|r|}{0.26} \\
Ours (K=5, $\lambda= 0.1$) & \textbf{49.80} & \textbf{89.94} & 87.27 & \textbf{88.31} & \textbf{73.84} & \textbf{65.34} & 85.17 & \textbf{70.83} & \textbf{76.27}\vspace{-0.2cm} \\
 & \VarCell{r}{0.69} & \VarCell{r}{0.14} & \VarCell{r}{0.27} & \VarCell{r}{0.08} & \VarCell{r}{0.19} & \VarCell{r}{0.32} & \VarCell{r}{0.11} & \VarCell{r}{0.38}\vspace{-0.1cm} & \VarCell{|r|}{0.12} \\
Ours (K=5, $\lambda= 0.5$) & 48.66 & 89.71 & 87.21 & 88.20 & 73.62 & 65.14 & 85.18 & 70.02 & 75.95\vspace{-0.2cm} \\
 & \VarCell{r}{0.43} & \VarCell{r}{0.11} & \VarCell{r}{0.36} & \VarCell{r}{0.09} & \VarCell{r}{0.16} & \VarCell{r}{0.28} & \VarCell{r}{0.15} & \VarCell{r}{0.33}\vspace{-0.1cm} & \VarCell{|r|}{0.10} \\
Ours (K=5, $\lambda= 1$) & 48.43 & 89.90 & 87.02 & 87.86 & 73.22 & 64.64 & 85.07 & 70.64 & 75.85\vspace{-0.2cm} \\
 & \VarCell{r}{0.83} & \VarCell{r}{0.17} & \VarCell{r}{0.39} & \VarCell{r}{0.08} & \VarCell{r}{0.18} & \VarCell{r}{0.83} & \VarCell{r}{0.15} & \VarCell{r}{0.40} & \VarCell{|r|}{0.17} \\
\hline
\end{tabular}
}
\centering
\caption{The scores on the GLUE development set. We compare different methods using BERT$_{\text{Base}}$ and BERT$_{\text{Large}}$ in GLUE 100 and 1k. }
\label{tb:GLUE_all_scores}
\end{table*}

\begin{table*}[t!]
\scalebox{0.9}{
\begin{tabular}{|c|cccccccc|c|}
\multicolumn{10}{c}{GLUE Full  (BERT Base)} \\ \hline
  & CoLA &  SST &  MRPC &  STS-B &  QQP &  MNLI &  QNLI &  RTE &  Avg.\\ 
  & 8.5k &  67k &  3.5k &  5.7k &  363k & 392k  &  108k & 2.5k & - \\
  & MCC  &  Acc &  F1   &  Spearman & F1 & Acc  &  Acc  &  Acc &    -  \\ \hline
MTL$\dagger$ & 49.4 & 91.2 & 89.1 & 88.3& 89.0 &82.0&90.5 & 70.8 & 81.4 \\ \hline
Pretrained & 59.09 & 92.71 & 89.82 & 88.13 & 87.29 & 84.33 & 91.11 & 64.42 & 82.05\vspace{-0.2cm} \\
 & \VarCell{r}{0.37} & \VarCell{r}{0.07} & \VarCell{r}{0.18} & \VarCell{r}{0.06} & \VarCell{r}{0.09} & \VarCell{r}{0.07} & \VarCell{r}{0.09} & \VarCell{r}{0.42}\vspace{-0.1cm} & \VarCell{|r|}{0.08} \\
MTL & \textbf{59.36} & 92.44 & 90.18 & 89.86 & \textbf{88.01} & 84.44 & 91.61 & 70.81 & 83.30\vspace{-0.2cm} \\
 & \VarCell{r}{0.28} & \VarCell{r}{0.06} & \VarCell{r}{0.14} & \VarCell{r}{0.04} & \VarCell{r}{0.04} & \VarCell{r}{0.05} & \VarCell{r}{0.04} & \VarCell{r}{0.46}\vspace{-0.1cm} & \VarCell{|r|}{0.07} \\
Ours (K=1) & 58.64 & \textbf{92.83} & 90.83 & 89.99 & 87.96 & \textbf{84.66} & 91.60 & 70.81 & 83.40\vspace{-0.2cm} \\
 & \VarCell{r}{0.40} & \VarCell{r}{0.06} & \VarCell{r}{0.12} & \VarCell{r}{0.05} & \VarCell{r}{0.06} & \VarCell{r}{0.07} & \VarCell{r}{0.05} & \VarCell{r}{0.32}\vspace{-0.1cm} & \VarCell{|r|}{0.07} \\
Ours (K=5, $\lambda= 0$) & 58.38 & 92.53 & 90.84 & 89.94 & 87.91 & 84.48 & 91.59 & 71.74 & 83.41\vspace{-0.2cm} \\
 & \VarCell{r}{0.34} & \VarCell{r}{0.09} & \VarCell{r}{0.14} & \VarCell{r}{0.04} & \VarCell{r}{0.04} & \VarCell{r}{0.06} & \VarCell{r}{0.06} & \VarCell{r}{0.38}\vspace{-0.1cm} & \VarCell{|r|}{0.07} \\
Ours (K=5, $\lambda= 0.1$) & 58.67 & 92.78 & 90.67 & \textbf{90.01} & 87.95 & 84.56 & 91.59 & 71.76 & \textbf{83.47}\vspace{-0.2cm} \\
 & \VarCell{r}{0.27} & \VarCell{r}{0.08} & \VarCell{r}{0.19} & \VarCell{r}{0.03} & \VarCell{r}{0.09} & \VarCell{r}{0.06} & \VarCell{r}{0.05} & \VarCell{r}{0.20}\vspace{-0.1cm} & \VarCell{|r|}{0.05} \\
Ours (K=5, $\lambda= 0.5$) & 59.01 & 92.70 & \textbf{90.96} & 89.99 & 87.86 & 84.62 & \textbf{91.66} & 71.14 & \textbf{83.47}\vspace{-0.2cm} \\
 & \VarCell{r}{0.22} & \VarCell{r}{0.08} & \VarCell{r}{0.17} & \VarCell{r}{0.04} & \VarCell{r}{0.07} & \VarCell{r}{0.08} & \VarCell{r}{0.07} & \VarCell{r}{0.51}\vspace{-0.1cm} & \VarCell{|r|}{0.08} \\
Ours (K=5, $\lambda= 1$) & 58.66 & 92.69 & 90.64 & 89.96 & 87.88 & 84.55 & 91.58 & \textbf{71.76} & 83.43\vspace{-0.2cm} \\
 & \VarCell{r}{0.28} & \VarCell{r}{0.08} & \VarCell{r}{0.20} & \VarCell{r}{0.02} & \VarCell{r}{0.10} & \VarCell{r}{0.07} & \VarCell{r}{0.06} & \VarCell{r}{0.35} & \VarCell{|r|}{0.07} \\
 \hline
\multicolumn{10}{c}{GLUE Fulll (BERT Large)} \\ \hline
  & CoLA &  SST &  MRPC &  STS-B &  QQP &  MNLI &  QNLI &  RTE &  Avg.\\ \hline
MTL & 62.42 & 93.94 & 90.93 & 90.10 & 86.26 & 84.53 & 92.45 & 72.49 & 84.13\vspace{-0.2cm} \\
 & \VarCell{r}{0.26} & \VarCell{r}{0.12} & \VarCell{r}{0.22} & \VarCell{r}{0.06} & \VarCell{r}{0.11} & \VarCell{r}{0.19} & \VarCell{r}{0.06} & \VarCell{r}{0.75}\vspace{-0.1cm} & \VarCell{|r|}{0.11} \\
Ours (K=1) & 62.62 & 93.82 & \textbf{91.26} & 89.89 & 86.36 & \textbf{85.09} & 92.56 & \textbf{75.17} & 84.59\vspace{-0.2cm} \\
 & \VarCell{r}{0.32} & \VarCell{r}{0.11} & \VarCell{r}{0.10} & \VarCell{r}{0.06} & \VarCell{r}{0.07} & \VarCell{r}{0.03} & \VarCell{r}{0.06} & \VarCell{r}{0.44}\vspace{-0.1cm} & \VarCell{|r|}{0.07} \\
Ours (K=5, $\lambda= 0$) & 62.81 & 93.93 & 90.69 & 90.04 & 86.36 & 84.84 & 92.53 & 74.96 & 84.51\vspace{-0.2cm} \\
 & \VarCell{r}{0.19} & \VarCell{r}{0.10} & \VarCell{r}{0.15} & \VarCell{r}{0.06} & \VarCell{r}{0.08} & \VarCell{r}{0.11} & \VarCell{r}{0.05} & \VarCell{r}{0.29}\vspace{-0.1cm} & \VarCell{|r|}{0.05} \\
Ours (K=5, $\lambda= 0.1$) & 62.63 & 93.86 & 91.03 & \textbf{90.25} & \textbf{86.42} & 84.96 & \textbf{92.59} & 75.16 & \textbf{84.61}\vspace{-0.2cm} \\
 & \VarCell{r}{0.36} & \VarCell{r}{0.08} & \VarCell{r}{0.15} & \VarCell{r}{0.05} & \VarCell{r}{0.06} & \VarCell{r}{0.09} & \VarCell{r}{0.05} & \VarCell{r}{0.45}\vspace{-0.1cm} & \VarCell{|r|}{0.08} \\
Ours (K=5, $\lambda= 0.5$) & 62.26 & \textbf{94.03} & 90.92 & 90.11 & 86.39 & 84.84 & 92.56 & 74.87 & 84.49\vspace{-0.2cm} \\
 & \VarCell{r}{0.34} & \VarCell{r}{0.05} & \VarCell{r}{0.11} & \VarCell{r}{0.05} & \VarCell{r}{0.07} & \VarCell{r}{0.12} & \VarCell{r}{0.07} & \VarCell{r}{0.49}\vspace{-0.1cm} & \VarCell{|r|}{0.08} \\
Ours (K=5, $\lambda= 1$) & \textbf{63.30} & 93.97 & 90.83 & 90.11 & 86.33 & 84.99 & 92.43 & 74.98 & 84.61\vspace{-0.2cm} \\
 & \VarCell{r}{0.23} & \VarCell{r}{0.09} & \VarCell{r}{0.18} & \VarCell{r}{0.05} & \VarCell{r}{0.13} & \VarCell{r}{0.10} & \VarCell{r}{0.06} & \VarCell{r}{0.46}& \VarCell{|r|}{0.07} \\
\hline

\end{tabular}
}
\centering
\caption{The scores on the GLUE development set. We compare different methods using BERT$_{\text{Base}}$ and BERT$_{\text{Large}}$ in GLUE Full. $\dagger$The number copied from \citet{aroca2020losses}. }
\label{tb:GLUE_all_scores_full}
\end{table*}

\begin{table*}[t!]
\scalebox{0.9}{
\begin{tabular}{|c|ccccccccc|c|}
\multicolumn{11}{c}{SuperGLUE 100 (BERT Base)} \\
\hline
  & BoolQ & \multicolumn{2}{c}{CB} & COPA & \multicolumn{2}{c}{MultiRC} & RTE & WiC & WSC & Avg.\\
  & Acc	& Acc & F1	& Acc& F1	& EM	& Acc	& Acc	& Acc & - \\ \hline
Pretrained & 61.21 & 77.68 & \textbf{74.53} & 59.63 & 53.81 & 1.27 & 54.41 & 55.78 & 60.16 & 57.18\vspace{-0.2cm} \\
 & \VarCell{r}{0.30} & \VarCell{r}{1.02} & \VarCell{r}{2.42} & \VarCell{r}{1.05} & \VarCell{r}{1.23} & \VarCell{r}{0.21} & \VarCell{r}{0.58} & \VarCell{r}{0.54} & \VarCell{r}{1.06}\vspace{-0.1cm} & \VarCell{|r|}{0.43} \\
MTL & 61.97 & 77.23 & 72.73 & 59.69 & 52.74 & 1.41 & 56.13 & 56.03 & 61.67 & 57.50\vspace{-0.2cm} \\
 & \VarCell{r}{0.11} & \VarCell{r}{1.27} & \VarCell{r}{2.24} & \VarCell{r}{1.09} & \VarCell{r}{1.16} & \VarCell{r}{0.24} & \VarCell{r}{1.10} & \VarCell{r}{0.39} & \VarCell{r}{0.62}\vspace{-0.1cm} & \VarCell{|r|}{0.41} \\
Ours (K=1) & 61.53 & 76.34 & 70.84 & 58.81 & 54.53 & 1.56 & 57.27 & 56.03 & 61.66 & 57.31\vspace{-0.2cm} \\
 & \VarCell{r}{0.20} & \VarCell{r}{1.27} & \VarCell{r}{1.97} & \VarCell{r}{0.68} & \VarCell{r}{0.83} & \VarCell{r}{0.18} & \VarCell{r}{0.94} & \VarCell{r}{0.45} & \VarCell{r}{0.53}\vspace{-0.1cm} & \VarCell{|r|}{0.35} \\
Ours (K=5, $\lambda= 0$) & 62.01 & \textbf{79.14} & 72.13 & 60.44 & 55.34 & 3.09 & \textbf{58.17} & 56.99 & 61.91 & 58.29\vspace{-0.2cm} \\
 & \VarCell{r}{0.15} & \VarCell{r}{0.73} & \VarCell{r}{1.82} & \VarCell{r}{0.57} & \VarCell{r}{0.47} & \VarCell{r}{0.41} & \VarCell{r}{1.09} & \VarCell{r}{0.42} & \VarCell{r}{0.55}\vspace{-0.1cm} & \VarCell{|r|}{0.33} \\
Ours (K=5, $\lambda= 0.1$) & 62.04 & 78.79 & 72.67 & 60.63 & 54.31 & 3.24 & 58.15 & 56.74 & 61.24 & 58.20\vspace{-0.2cm} \\
 & \VarCell{r}{0.16} & \VarCell{r}{1.06} & \VarCell{r}{1.34} & \VarCell{r}{0.65} & \VarCell{r}{0.74} & \VarCell{r}{0.39} & \VarCell{r}{1.35} & \VarCell{r}{0.38} & \VarCell{r}{0.57}\vspace{-0.1cm} & \VarCell{|r|}{0.31} \\
Ours (K=5, $\lambda= 0.5$) & \textbf{62.09} & 78.03 & 71.98 & \textbf{61.19} & 55.72 & 3.33 & 57.60 & \textbf{57.54} & 61.78 & \textbf{58.41}\vspace{-0.2cm} \\
 & \VarCell{r}{0.14} & \VarCell{r}{0.95} & \VarCell{r}{1.83} & \VarCell{r}{1.13} & \VarCell{r}{0.76} & \VarCell{r}{0.40} & \VarCell{r}{1.14} & \VarCell{r}{0.60} & \VarCell{r}{0.61}\vspace{-0.1cm} & \VarCell{|r|}{0.38} \\
Ours (K=5, $\lambda= 1$) & 61.94 & 77.80 & 69.21 & 59.94 & \textbf{55.96} & \textbf{3.97} & 57.76 & 56.29 & \textbf{62.57} & 57.84\vspace{-0.2cm} \\
 & \VarCell{r}{0.25} & \VarCell{r}{0.77} & \VarCell{r}{2.39} & \VarCell{r}{0.59} & \VarCell{r}{0.45} & \VarCell{r}{0.32} & \VarCell{r}{1.17} & \VarCell{r}{0.42} & \VarCell{r}{0.32} & \VarCell{|r|}{0.40} \\
 \hline
\multicolumn{11}{c}{SuperGLUE 100 (BERT Large)} \\ \hline
& BoolQ & \multicolumn{2}{c}{CB} & COPA & \multicolumn{2}{c}{MultiRC} & RTE & WiC & WSC & Avg.\\ \hline
MTL & 62.03 & 78.14 & 74.21 & 64.31 & \textbf{55.76} & 1.32 & 58.24 & 56.42 & \textbf{62.28} & 59.03\vspace{-0.2cm} \\
 & \VarCell{r}{0.13} & \VarCell{r}{1.80} & \VarCell{r}{3.23} & \VarCell{r}{1.04} & \VarCell{r}{1.40} & \VarCell{r}{0.29} & \VarCell{r}{1.34} & \VarCell{r}{0.35} & \VarCell{r}{0.63}\vspace{-0.1cm} & \VarCell{|r|}{0.54} \\
Ours (K=1) & 60.49 & 77.36 & 71.39 & 61.63 & 52.96 & 1.18 & 57.04 & 55.79 & 61.43 & 57.35\vspace{-0.2cm} \\
 & \VarCell{r}{0.38} & \VarCell{r}{0.92} & \VarCell{r}{2.38} & \VarCell{r}{1.13} & \VarCell{r}{1.04} & \VarCell{r}{0.18} & \VarCell{r}{0.74} & \VarCell{r}{0.42} & \VarCell{r}{0.74}\vspace{-0.1cm} & \VarCell{|r|}{0.42} \\
Ours (K=5, $\lambda= 0$) & 62.08 & 78.90 & 75.26 & \textbf{64.63} & 51.08 & 3.46 & 61.07 & \textbf{57.38} & 61.37 & 59.46\vspace{-0.2cm} \\
 & \VarCell{r}{0.07} & \VarCell{r}{1.45} & \VarCell{r}{2.40} & \VarCell{r}{1.13} & \VarCell{r}{1.23} & \VarCell{r}{0.37} & \VarCell{r}{0.58} & \VarCell{r}{0.69} & \VarCell{r}{1.07}\vspace{-0.1cm} & \VarCell{|r|}{0.44} \\
Ours (K=5, $\lambda= 0.1$) & 62.18 & 80.36 & 77.08 & 64.19 & 51.48 & \textbf{3.61} & \textbf{62.43} & 57.37 & 61.13 & \textbf{59.88}\vspace{-0.2cm} \\
 & \VarCell{r}{0.01} & \VarCell{r}{1.46} & \VarCell{r}{2.46} & \VarCell{r}{0.93} & \VarCell{r}{1.47} & \VarCell{r}{0.35} & \VarCell{r}{0.47} & \VarCell{r}{0.77} & \VarCell{r}{0.71}\vspace{-0.1cm} & \VarCell{|r|}{0.43} \\
Ours (K=5, $\lambda= 0.5$) & \textbf{62.19} & \textbf{80.69} & \textbf{77.28} & 63.25 & 52.87 & 2.99 & 60.07 & 56.80 & 61.24 & 59.42\vspace{-0.2cm} \\
 & \VarCell{r}{0.01} & \VarCell{r}{0.95} & \VarCell{r}{1.79} & \VarCell{r}{0.89} & \VarCell{r}{1.00} & \VarCell{r}{0.35} & \VarCell{r}{0.77} & \VarCell{r}{0.57} & \VarCell{r}{0.76}\vspace{-0.1cm} & \VarCell{|r|}{0.34} \\
Ours (K=5, $\lambda= 1$) & 62.06 & 79.14 & 73.22 & 62.69 & 50.01 & 3.14 & 60.81 & 57.24 & 61.44 & 58.74\vspace{-0.2cm} \\
 & \VarCell{r}{0.08} & \VarCell{r}{1.47} & \VarCell{r}{3.05} & \VarCell{r}{0.88} & \VarCell{r}{1.54} & \VarCell{r}{0.41} & \VarCell{r}{1.00} & \VarCell{r}{0.49} & \VarCell{r}{0.62} & \VarCell{|r|}{0.50} \\
 \hline
\multicolumn{11}{c}{SuperGLUE 1k (BERT Base)} \\ \hline
& BoolQ & \multicolumn{2}{c}{CB} & COPA & \multicolumn{2}{c}{MultiRC} & RTE & WiC & WSC & Avg.\\ \hline
 Pretrained & 62.89 & \textbf{87.00} & \textbf{85.63} & 60.94 & 55.37 & 5.19 & 59.39 & 60.40 & 64.54 & 61.55\vspace{-0.2cm} \\
 & \VarCell{r}{0.27} & \VarCell{r}{0.80} & \VarCell{r}{1.51} & \VarCell{r}{0.53} & \VarCell{r}{0.81} & \VarCell{r}{0.72} & \VarCell{r}{0.56} & \VarCell{r}{0.44} & \VarCell{r}{0.34}\vspace{-0.1cm} & \VarCell{|r|}{0.37} \\
MTL & 63.38 & 85.49 & 82.85 & 60.91 & 56.52 & 7.44 & 65.96 & 63.61 & \textbf{65.61} & 62.94\vspace{-0.2cm} \\
 & \VarCell{r}{0.39} & \VarCell{r}{0.79} & \VarCell{r}{1.37} & \VarCell{r}{0.52} & \VarCell{r}{0.69} & \VarCell{r}{0.64} & \VarCell{r}{1.06} & \VarCell{r}{0.21} & \VarCell{r}{0.32}\vspace{-0.1cm} & \VarCell{|r|}{0.36} \\
Ours (K=1) & \textbf{63.87} & 86.83 & 84.28 & 60.63 & 58.68 & 7.86 & 66.34 & 65.00 & 64.09 & 63.35\vspace{-0.2cm} \\
 & \VarCell{r}{0.45} & \VarCell{r}{0.47} & \VarCell{r}{0.65} & \VarCell{r}{0.39} & \VarCell{r}{0.19} & \VarCell{r}{0.14} & \VarCell{r}{0.38} & \VarCell{r}{0.29} & \VarCell{r}{0.35}\vspace{-0.1cm} & \VarCell{|r|}{0.18} \\
Ours (K=5, $\lambda= 0$) & 63.27 & 86.49 & 82.18 & \textbf{62.88} & 59.03 & 7.76 & \textbf{67.60} & 65.18 & 65.03 & 63.71\vspace{-0.2cm} \\
 & \VarCell{r}{0.41} & \VarCell{r}{0.46} & \VarCell{r}{0.90} & \VarCell{r}{0.64} & \VarCell{r}{0.72} & \VarCell{r}{0.45} & \VarCell{r}{0.48} & \VarCell{r}{0.34} & \VarCell{r}{0.37}\vspace{-0.1cm} & \VarCell{|r|}{0.26} \\
Ours (K=5, $\lambda= 0.1$) & 63.20 & 86.38 & 82.63 & 62.53 & 59.16 & \textbf{8.36} & 67.11 & 65.11 & 64.45 & 63.61\vspace{-0.2cm} \\
 & \VarCell{r}{0.39} & \VarCell{r}{0.57} & \VarCell{r}{0.98} & \VarCell{r}{0.61} & \VarCell{r}{0.34} & \VarCell{r}{0.23} & \VarCell{r}{0.58} & \VarCell{r}{0.27} & \VarCell{r}{0.38}\vspace{-0.1cm} & \VarCell{|r|}{0.27} \\
Ours (K=5, $\lambda= 0.5$) & 63.25 & 86.83 & 84.14 & 61.97 & \textbf{59.57} & 8.10 & 66.71 & \textbf{65.38} & 64.78 & \textbf{63.78}\vspace{-0.2cm} \\
 & \VarCell{r}{0.38} & \VarCell{r}{0.57} & \VarCell{r}{0.91} & \VarCell{r}{0.51} & \VarCell{r}{0.20} & \VarCell{r}{0.41} & \VarCell{r}{0.49} & \VarCell{r}{0.42} & \VarCell{r}{0.42}\vspace{-0.1cm} & \VarCell{|r|}{0.25} \\
Ours (K=5, $\lambda= 1$) & 63.12 & 86.88 & 83.97 & 61.66 & 58.57 & 7.70 & 66.83 & 65.15 & 64.56 & 63.56\vspace{-0.2cm} \\
 & \VarCell{r}{0.42} & \VarCell{r}{0.51} & \VarCell{r}{0.78} & \VarCell{r}{0.52} & \VarCell{r}{0.46} & \VarCell{r}{0.22} & \VarCell{r}{0.41} & \VarCell{r}{0.39} & \VarCell{r}{0.36} & \VarCell{|r|}{0.22} \\
 \hline
\multicolumn{11}{c}{SuperGLUE 1k (BERT Large)} \\ \hline
& BoolQ & \multicolumn{2}{c}{CB} & COPA & \multicolumn{2}{c}{MultiRC} & RTE & WiC & WSC & Avg.\\ \hline
MTL & 63.86 & \textbf{88.67} & \textbf{87.83} & 67.22 & 56.56 & 7.52 & 68.68 & 66.16 & 64.67 & 65.21\vspace{-0.2cm} \\
 & \VarCell{r}{0.42} & \VarCell{r}{0.87} & \VarCell{r}{1.46} & \VarCell{r}{0.73} & \VarCell{r}{0.64} & \VarCell{r}{0.53} & \VarCell{r}{0.83} & \VarCell{r}{0.29} & \VarCell{r}{0.29}\vspace{-0.1cm} & \VarCell{|r|}{0.38} \\
Ours (K=1) & 63.22 & 87.11 & 85.15 & 66.09 & \textbf{59.81} & 6.86 & 67.89 & 65.45 & \textbf{65.28} & 64.67\vspace{-0.2cm} \\
 & \VarCell{r}{0.35} & \VarCell{r}{0.85} & \VarCell{r}{1.74} & \VarCell{r}{0.63} & \VarCell{r}{0.35} & \VarCell{r}{0.87} & \VarCell{r}{0.77} & \VarCell{r}{0.23} & \VarCell{r}{0.49}\vspace{-0.1cm} & \VarCell{|r|}{0.43} \\
Ours (K=5, $\lambda= 0$) & 63.73 & 87.61 & 86.19 & \textbf{70.12} & 54.34 & 7.39 & 69.09 & 66.84 & 64.85 & 65.43\vspace{-0.2cm} \\
 & \VarCell{r}{0.47} & \VarCell{r}{0.75} & \VarCell{r}{1.11} & \VarCell{r}{0.66} & \VarCell{r}{3.53} & \VarCell{r}{0.79} & \VarCell{r}{0.61} & \VarCell{r}{0.33} & \VarCell{r}{0.33}\vspace{-0.1cm} & \VarCell{|r|}{0.38} \\
Ours (K=5, $\lambda= 0.1$) & \textbf{64.73} & 87.51 & 87.14 & 68.09 & 58.56 & \textbf{8.96} & 68.85 & 66.57 & 64.31 & 65.59\vspace{-0.2cm} \\
 & \VarCell{r}{0.52} & \VarCell{r}{0.60} & \VarCell{r}{0.82} & \VarCell{r}{0.65} & \VarCell{r}{0.34} & \VarCell{r}{0.20} & \VarCell{r}{0.53} & \VarCell{r}{0.43} & \VarCell{r}{0.46}\vspace{-0.1cm} & \VarCell{|r|}{0.25} \\
Ours (K=5, $\lambda= 0.5$) & 63.55 & 87.83 & 87.45 & 68.88 & 58.66 & 8.86 & \textbf{69.78} & 66.64 & 64.44 & \textbf{65.84}\vspace{-0.2cm} \\
 & \VarCell{r}{0.46} & \VarCell{r}{0.52} & \VarCell{r}{0.74} & \VarCell{r}{0.78} & \VarCell{r}{0.37} & \VarCell{r}{0.21} & \VarCell{r}{0.45} & \VarCell{r}{0.36} & \VarCell{r}{0.36}\vspace{-0.1cm} & \VarCell{|r|}{0.25} \\
Ours (K=5, $\lambda= 1$) & 63.83 & 86.72 & 84.79 & 67.75 & 56.87 & 8.14 & 68.33 & \textbf{66.93} & 64.84 & 65.00\vspace{-0.2cm} \\
 & \VarCell{r}{0.38} & \VarCell{r}{0.68} & \VarCell{r}{1.06} & \VarCell{r}{0.59} & \VarCell{r}{0.76} & \VarCell{r}{0.61} & \VarCell{r}{0.44} & \VarCell{r}{0.30} & \VarCell{r}{0.38}\vspace{-0.1cm} & \VarCell{|r|}{0.29} \\
\hline
\end{tabular}
}
\centering
\caption{The scores on the development set of the tasks in SuperGLUE except for ReCoRD. We compare different methods using BERT$_{\text{Base}}$ and BERT$_{\text{Large}}$ in SuperGLUE 100 and 1k. }
\label{tb:SuperGLUE_all_scores}
\end{table*}

\begin{table*}[t!]
\scalebox{0.8}{
\begin{tabular}{|c|ccccccccccc|c|}
\multicolumn{13}{c}{SuperGLUE Full (BERT Base)} \\ \hline
  & BoolQ & \multicolumn{2}{c}{CB} & COPA & \multicolumn{2}{c}{MultiRC} & RTE & WiC & WSC & \multicolumn{2}{c|}{ReCoRD} & Avg.\\
  & 9.4k & \multicolumn{2}{c}{250} & 400 & \multicolumn{2}{c}{5.1k} & 2.5k & 6k & 554 & \multicolumn{2}{c|}{101k} & - \\
  & Acc & Acc & F1 & Acc & F1 & EM & Acc & Acc & Acc & F1 & EM & - \\
Pretrained & 74.01 & \textbf{87.00} & \textbf{85.63} & 60.94 & \textbf{65.93} & \textbf{16.72} & 65.76 & 66.85 & 64.33 & 58.78 & 58.10 & 65.04\vspace{-0.2cm} \\
 & \VarCell{r}{0.34} & \VarCell{r}{0.80} & \VarCell{r}{1.51} & \VarCell{r}{0.53} & \VarCell{r}{0.13} & \VarCell{r}{0.17} & \VarCell{r}{0.44} & \VarCell{r}{0.29} & \VarCell{r}{0.40} & \VarCell{r}{0.62} & \VarCell{r}{0.62}\vspace{-0.1cm} & \VarCell{|r|}{0.36} \\
MTL & 77.46 & 85.49 & 82.85 & 60.91 & 65.45 & 16.03 & 72.09 & 69.77 & 65.56 & 59.10 & 58.38 & 66.33\vspace{-0.2cm} \\
 & \VarCell{r}{0.24} & \VarCell{r}{0.79} & \VarCell{r}{1.37} & \VarCell{r}{0.52} & \VarCell{r}{0.13} & \VarCell{r}{0.15} & \VarCell{r}{0.59} & \VarCell{r}{0.25} & \VarCell{r}{0.32} & \VarCell{r}{0.39} & \VarCell{r}{0.40}\vspace{-0.1cm} & \VarCell{|r|}{0.33} \\
Ours (K=1) & 77.46 & 86.83 & 84.28 & 60.63 & 65.34 & 15.89 & 72.19 & 70.76 & 64.55 & 57.68 & 56.98 & 66.29\vspace{-0.2cm} \\
 & \VarCell{r}{0.13} & \VarCell{r}{0.47} & \VarCell{r}{0.65} & \VarCell{r}{0.39} & \VarCell{r}{0.19} & \VarCell{r}{0.18} & \VarCell{r}{0.55} & \VarCell{r}{0.16} & \VarCell{r}{0.32} & \VarCell{r}{0.97} & \VarCell{r}{0.96}\vspace{-0.1cm} & \VarCell{|r|}{0.18} \\
Ours (K=5, $\lambda= 0$) & \textbf{77.57} & 86.49 & 82.18 & \textbf{62.87} & 65.79 & 16.03 & \textbf{72.77} & 70.68 & 65.14 & 60.20 & 59.48 & 66.80\vspace{-0.2cm} \\
 & \VarCell{r}{0.31} & \VarCell{r}{0.46} & \VarCell{r}{0.90} & \VarCell{r}{0.64} & \VarCell{r}{0.11} & \VarCell{r}{0.20} & \VarCell{r}{0.44} & \VarCell{r}{0.21} & \VarCell{r}{0.26} & \VarCell{r}{0.57} & \VarCell{r}{0.56}\vspace{-0.1cm} & \VarCell{|r|}{0.25} \\
Ours (K=5, $\lambda= 0.1$) & 77.29 & 86.38 & 82.63 & 62.53 & 65.66 & 16.17 & 72.24 & 70.58 & 65.31 & \textbf{60.27} & \textbf{59.55} & 66.74\vspace{-0.2cm} \\
 & \VarCell{r}{0.16} & \VarCell{r}{0.57} & \VarCell{r}{0.98} & \VarCell{r}{0.61} & \VarCell{r}{0.13} & \VarCell{r}{0.24} & \VarCell{r}{0.59} & \VarCell{r}{0.18} & \VarCell{r}{0.28} & \VarCell{r}{0.48} & \VarCell{r}{0.48}\vspace{-0.1cm} & \VarCell{|r|}{0.26} \\
Ours (K=5, $\lambda= 0.5$) & 76.84 & 86.83 & 84.14 & 61.97 & 65.58 & 15.84 & 72.11 & \textbf{70.88} & \textbf{65.81} & 59.85 & 59.10 & \textbf{66.80}\vspace{-0.2cm} \\
 & \VarCell{r}{0.27} & \VarCell{r}{0.57} & \VarCell{r}{0.91} & \VarCell{r}{0.51} & \VarCell{r}{0.13} & \VarCell{r}{0.22} & \VarCell{r}{0.39} & \VarCell{r}{0.20} & \VarCell{r}{0.40} & \VarCell{r}{0.48} & \VarCell{r}{0.48}\vspace{-0.1cm} & \VarCell{|r|}{0.24} \\
Ours (K=5, $\lambda= 1$) & 76.69 & 86.88 & 83.97 & 61.66 & 65.33 & 16.23 & 71.54 & 70.43 & 65.14 & 58.62 & 57.88 & 66.39\vspace{-0.2cm} \\
 & \VarCell{r}{0.27} & \VarCell{r}{0.51} & \VarCell{r}{0.78} & \VarCell{r}{0.52} & \VarCell{r}{0.20} & \VarCell{r}{0.18} & \VarCell{r}{0.68} & \VarCell{r}{0.32} & \VarCell{r}{0.30} & \VarCell{r}{0.97} & \VarCell{r}{0.95} & \VarCell{|r|}{0.22} \\ \hline
\multicolumn{13}{c}{SuperGLUE Full (BERT Large)} \\ \hline
MTL & 77.78 & \textbf{88.67} & \textbf{87.83} & 67.22 & 65.93 & 16.68 & 71.97 & 71.08 & 64.37 & 69.60 & 68.85 & 69.16\vspace{-0.2cm} \\
 & \VarCell{r}{0.35} & \VarCell{r}{0.87} & \VarCell{r}{1.46} & \VarCell{r}{0.73} & \VarCell{r}{0.19} & \VarCell{r}{0.29} & \VarCell{r}{1.08} & \VarCell{r}{0.22} & \VarCell{r}{0.24} & \VarCell{r}{0.60} & \VarCell{r}{0.61}\vspace{-0.1cm} & \VarCell{|r|}{0.37} \\
Ours (K=1) & 78.04 & 87.11 & 85.15 & 66.09 & 65.96 & 16.49 & \textbf{75.54} & 70.62 & 65.02 & \textbf{70.12} & \textbf{69.43} & 69.24\vspace{-0.2cm} \\
 & \VarCell{r}{0.40} & \VarCell{r}{0.85} & \VarCell{r}{1.74} & \VarCell{r}{0.63} & \VarCell{r}{0.15} & \VarCell{r}{0.28} & \VarCell{r}{0.43} & \VarCell{r}{0.14} & \VarCell{r}{0.36} & \VarCell{r}{0.10} & \VarCell{r}{0.10}\vspace{-0.1cm} & \VarCell{|r|}{0.41} \\
Ours (K=5, $\lambda= 0$) & 78.21 & 87.61 & 86.19 & \textbf{70.12} & 64.91 & 15.62 & 73.34 & \textbf{71.73} & 65.26 & 69.28 & 68.58 & 69.56\vspace{-0.2cm} \\
 & \VarCell{r}{0.25} & \VarCell{r}{0.75} & \VarCell{r}{1.11} & \VarCell{r}{0.66} & \VarCell{r}{0.71} & \VarCell{r}{1.18} & \VarCell{r}{0.58} & \VarCell{r}{0.20} & \VarCell{r}{0.38} & \VarCell{r}{0.45} & \VarCell{r}{0.47}\vspace{-0.1cm} & \VarCell{|r|}{0.31} \\
Ours (K=5, $\lambda= 0.1$) & \textbf{78.70} & 87.51 & 87.14 & 68.09 & 65.83 & \textbf{17.67} & 75.36 & 71.41 & \textbf{65.44} & 69.88 & 69.25 & \textbf{69.98}\vspace{-0.2cm} \\
 & \VarCell{r}{0.20} & \VarCell{r}{0.60} & \VarCell{r}{0.82} & \VarCell{r}{0.65} & \VarCell{r}{0.13} & \VarCell{r}{0.25} & \VarCell{r}{0.32} & \VarCell{r}{0.24} & \VarCell{r}{0.52} & \VarCell{r}{0.35} & \VarCell{r}{0.40}\vspace{-0.1cm} & \VarCell{|r|}{0.24} \\
Ours (K=5, $\lambda= 0.5$) & 78.54 & 87.83 & 87.45 & 68.88 & \textbf{66.06} & 16.66 & 74.83 & 71.49 & 64.83 & 68.87 & 68.15 & 69.79\vspace{-0.2cm} \\
 & \VarCell{r}{0.39} & \VarCell{r}{0.52} & \VarCell{r}{0.74} & \VarCell{r}{0.78} & \VarCell{r}{0.16} & \VarCell{r}{0.33} & \VarCell{r}{0.62} & \VarCell{r}{0.20} & \VarCell{r}{0.36} & \VarCell{r}{0.52} & \VarCell{r}{0.51}\vspace{-0.1cm} & \VarCell{|r|}{0.25} \\
Ours (K=5, $\lambda= 1$) & 77.49 & 86.72 & 84.79 & 67.75 & 65.77 & 17.09 & 74.46 & 70.97 & 64.79 & 68.45 & 67.73 & 69.04\vspace{-0.2cm} \\
 & \VarCell{r}{0.29} & \VarCell{r}{0.68} & \VarCell{r}{1.06} & \VarCell{r}{0.59} & \VarCell{r}{0.27} & \VarCell{r}{0.41} & \VarCell{r}{0.47} & \VarCell{r}{0.21} & \VarCell{r}{0.40} & \VarCell{r}{0.37} & \VarCell{r}{0.39}\vspace{-0.1cm} & \VarCell{|r|}{0.27} \\ \hline
\end{tabular}
}
\centering
\caption{The scores on the SuperGLUE development set. We compare different methods using BERT$_{\text{Base}}$ and BERT$_{\text{Large}}$ in SuperGLUE Full.}
\label{tb:SuperGLUE_all_scores_full}
\end{table*}


\begin{table*}[t!]
\scalebox{0.9}{ 
\begin{tabular}{|c|ccccccc|c|}
\multicolumn{9}{c}{GLUE 100 ECE (BERT Base)} \\ \hline
& CoLA &  SST &  MRPC &    QQP &  MNLI &  QNLI &  RTE &  Avg.\\ \hline
Ours (K=1) & 27.15 & 19.96 & 10.90 & 32.18 & 32.70 & 23.37 & 30.26 & 25.22 \\
Ours (K=5, $\lambda=0.1$) & 24.21 & 14.40 & 20.06 & 16.01 & 17.02 & 6.53 & 9.96 & 15.46 \\ \hline
\multicolumn{9}{c}{GLUE 1k ECE (BERT Base)} \\ \hline
& CoLA &  SST &  MRPC &  QQP &  MNLI &  QNLI &  RTE &  Avg.\\ \hline
Ours (K=1) & 21.72 & 10.19 & 15.21 &  13.93 & 35.21 & 15.58 & 23.43 & 19.32  \\
Ours (K=5, $\lambda=0.1$) & 20.50 & 7.52 & 16.11 & 14.36 & 32.67 & 15.00 & 12.88 & 17.01 \\ \hline
\multicolumn{9}{c}{GLUE Full ECE (BERT Base)} \\ \hline
& CoLA &  SST &  MRPC &  QQP &  MNLI &  QNLI &  RTE &  Avg.\\ \hline
Ours (K=1) & 14.90 & 3.07 & 10.45 & 5.28 & 4.67 & 2.38 & 22.20 & 8.99 \\
Ours (K=5, $\lambda=0.1$) & 15.06 & 4.23 & 5.97 & 4.85 & 4.43 & 3.52 & 23.35 & 8.77 \\ \hline
\end{tabular}
}
\centering
\caption{The comparison of expected calibration error (ECE) in the classification tasks of GLUE.}
\label{tb:ECE_all_scores}
\end{table*}

\begin{table*}[t!]
\scalebox{1}{
\begin{tabular}{|c|ccccccc|c|}
\multicolumn{9}{c}{GLUE 100 (BERT Base)} \\ \hline
& CoLA &  SST &  MRPC &  QQP &  MNLI &  QNLI &  RTE &  Avg.\\ \hline
CLS vs ENS &  32.93 & 47.13 & 56.79 & 25.70 & 21.70 &	21.47$\dagger$ &	22.27$\dagger$ &	32.57 \\
Dropout vs ENS & 33.65 & 47.56 &	54.63 &	30.88 &	24.76 &	38.28 &	30.45$\dagger$ &	37.17 \\
Least vs ENS &  38.70 &	48.28 &	61.11 &	26.84 &	24.66 &	42.40 &	35.00 &	39.57 \\
ENS vs ENS & 35.34 & 42.53 & 59.88 & 31.87 & 26.58 &	43.13 &	31.36$\dagger$ &	38.67 \\ \hline

\multicolumn{9}{c}{GLUE 1k (BERT Base)} \\ \hline
& CoLA &  SST &  MRPC &  QQP &  MNLI &  QNLI &  RTE &  Avg.\\ \hline
CLS vs ENS & 53.25 & 46.98 & 46.91 & 34.76 & 32.63 &	49.45 &	25.45$\dagger$ & 41.35 \\
Dropout vs ENS & 45.55 & 54.31 & 46.30 & 49.91 & 37.76 &	53.50 &	31.36$\dagger$ & 45.53 \\
Least vs ENS & 59.01 &	60.92 &	51.54 &	48.24 &	37.68 &	54.53 &	30.00$\dagger$ & 48.85\\ 
ENS vs ENS & 57.09 & 59.48 & 50.31 &	50.62 &	41.00 &	56.14 &	36.36 & 50.14 \\ \hline
\end{tabular}
}
\centering
\caption{The overlapping ratio of the top 20\% most uncertain examples using different uncertainty estimation methods. All the scores are significantly larger than the random ratio 0.2 with p value < $10^{-4}$ except for the values beside $\dagger$.}
\label{tb:top_20_all}
\end{table*}

\end{document}